\documentclass[letterpaper]{sig-alternate-2013}
\usepackage[utf8]{inputenc}
\usepackage{balance}
\usepackage{times}

\usepackage{amssymb,amsmath,mathrsfs}
\usepackage{graphicx}
\usepackage[linesnumbered,ruled,lined]{algorithm2e}
\usepackage[font=small,labelfont=bf]{caption} %,labelfont=bf,textfont=bf
\usepackage[small]{subfigure}
\usepackage{color}
\usepackage{url}
\usepackage{float}
\usepackage{multirow}
\usepackage{tabularx}
\usepackage{bbm}
\usepackage{enumitem}

\usepackage[table]{xcolor}

\newtheorem{definition}{Definition}

\newtheorem{hypothesis}{Hypothesis}

\newcommand{\nop}[1]{}
\def\ssp{\hspace*{0.4ex}}

\DeclareMathOperator{\RR}{\mathbb{R}}
\DeclareMathOperator{\EE}{\mathbb{E}}

\def\e{\mathrm{exp}}
\def\log{\mathrm{log}}

\def\ie{{\sl i.e.}}
\def\eg{{\sl e.g.}}

\def\L{\mathcal{L}}
\def\O{\mathcal{O}}
\def\F{\mathcal{F}}

\DeclareMathOperator{\R}{\mathcal{R}}
\DeclareMathOperator{\Rn}{\overline{\mathcal{R}}}
\def\I{\mathcal{I}}
\def\Z{\mathcal{Z}}
\def\M{\mathcal{M}}
\def\Y{\mathcal{Y}}
\def\Yn{\overline{\mathcal{Y}}}
\def\D{\mathcal{D}}
\def\P{\mathcal{P}}

\def\E{\mathcal{E}}
\def\T{\mathcal{T}}

\def\bu{\mathbf{u}}
\def\bv{\mathbf{v}}

\def\bc{\mathbf{c}}

\def\bz{\mathbf{z}}
\def\br{\mathbf{r}}

\def\bm{\mathbf{m}}
\def\by{\mathbf{y}}

\newcolumntype{x}{>{\hsize=.8\hsize}X}
\newcolumntype{a}{>{\hsize=.4\hsize}X}
\newcolumntype{b}{>{\hsize=1.6\hsize}X}

\newfont{\mycrnotice}{ptmr8t at 7pt}
\newfont{\myconfname}{ptmri8t at 7pt}
\let\crnotice\mycrnotice
\let\confname\myconfname
\permission{\copyright 2017 International World Wide Web Conference Committee \\ (IW3C2), published under Creative Commons CC BY 4.0 License.}
\conferenceinfo{WWW 2017,}{April 3--7, 2017, Perth, Australia.}
\copyrightetc{ACM \the\acmcopyr}
\crdata{978-1-4503-4913-0/17/04. \\
http://dx.doi.org/10.1145/3038912.3052708 \\
\includegraphics{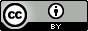}
}

\begin{document}

\title{CoType: Joint Extraction of Typed Entities and Relations with Knowledge Bases}

\author{
\alignauthor
$\quad$ Xiang Ren$^{\dag}\quad$ Zeqiu Wu$^{\dag}\quad$ Wenqi He$^{\dag}\quad$ Meng Qu$^{\dag}$ $\quad$  Clare R. Voss$^{\ddagger}\quad$ \\[0.3ex] Heng Ji$^{\sharp}\quad$ Tarek F. Abdelzaher$^{\dag}\quad$  Jiawei Han$^{\dag}$\\[0.5ex]
\affaddr{$^{\dag}$ University of Illinois at Urbana-Champaign, Urbana, IL, USA}\\
\affaddr{$^{\ddagger}$ Computational \&  Information Sciences Directorate, Army Research Laboratory, Adelphi, MD, USA}\\
\affaddr{$^{\sharp}$ Computer Science Department, Rensselaer Polytechnic Institute, USA}\\
\email{$^{\dag}$\{xren7, zeqiuwu1, wenqihe3, mengqu2, zaher, hanj\}@illinois.edu}
\email{$^{\ddagger}$clare.r.voss.civ@mail.mil$\ssp$ $^{\sharp}$jih@rpi.edu}
}

\maketitle

\begin{abstract}
Extracting entities and relations for types of interest from text is important for understanding massive text corpora. Traditionally, systems of entity relation extraction have relied on human-annotated corpora for training and adopted an \textit{incremental} pipeline. Such systems require additional human expertise to be ported to a new domain, and are vulnerable to errors cascading down the pipeline. In this paper, we investigate \textit{joint extraction of typed entities and relations} with labeled data \textit{heuristically} obtained from knowledge bases (\ie, distant supervision). As our algorithm for type labeling via distant supervision is \textit{context-agnostic}, noisy training data poses unique challenges for the task. We propose a novel \textit{domain-independent} framework, called \textsc{CoType}, that runs a data-driven text segmentation algorithm to extract entity mentions, and jointly embeds entity mentions, relation mentions, text features and type labels into two low-dimensional spaces (for entity and relation mentions respectively), where, in each space, objects whose types are close will also have similar representations. \textsc{CoType}, then using these learned embeddings, estimates the types of test (unlinkable) mentions. We formulate a joint optimization problem to learn embeddings from text corpora and knowledge bases, adopting a novel partial-label loss function for noisy labeled data and introducing an object "translation" function to capture the cross-constraints of entities and relations on each other. Experiments on three public datasets demonstrate the effectiveness  of \textsc{CoType} across different domains (\eg, news, biomedical), with an average of 25\% improvement in F1 score compared to the next best method.
\end{abstract}
%% problem
%In this paper, we investigate \textit{joint extraction of typed entities and relations with distant supervision}, to be the automatic detection of entity and relation mentions in text and jointly classifying them into a given set of types, with labeled data obtained from knowledge bases. 
%% challenges

%%%%%%%%%%%%%%%%%%%%%%%
% Introduction
%%%%%%%%%%%%%%%%%%%%%%%
\section{Introduction}
\label{sec:intro}

%% joint extraction of typed entities and relations is useful: background & examples
The extraction of entities and their relations is critical to understanding massive text corpora. Identifying the token spans in text that constitute entity mentions and assigning types (\eg, \begin{small}\texttt{person}, \texttt{company}\end{small}) to these spans as well as to the relations between entity mentions (\eg, \begin{small}\texttt{employed\_by}\end{small}) are key to structuring content from text corpora for further analytics. 
For example, when an extraction system finds a \begin{small}``\texttt{produce}"\end{small} relation between \begin{small}``\texttt{company}"\end{small} and \begin{small}``\texttt{product}"\end{small} entities in news articles, it supports answering questions like ``\textit{what products does company X produce?}".
%similarly, mining \texttt{treat} relations between \texttt{drug} and \texttt{disease} entities from biomedical papers helps answer questions like ``\textit{what drugs are used to treat a disease}". 
%
Once extracted, such structured information is used in many ways, \eg, as primitives in information extraction, knowledge base population~\cite{dong2014knowledge,west2014knowledge}, and question-answering systems~\cite{sun2015open,bian2008finding}.
Traditional systems for relation extraction~\cite{bach2007review,culotta2004dependency,guodong2005exploring} partition the process into several subtasks and solve them incrementally (\ie, detecting entities from text, labeling their types and then extracting their relations). Such systems treat the subtasks independently and so may propagate errors across subtasks in the process. Recent studies~\cite{li2014incremental,miwa2014modeling,roth2007global} focus on joint extraction methods to capture the inhereent linguistic dependencies between relations and entity arguments (\eg, the types of entity arguments help determine their relation type, and vice versa) to resolve error propagation.

\begin{figure}
\centering
\begin{small}
%\vspace{-0.0cm}
\includegraphics[width = 86 mm]{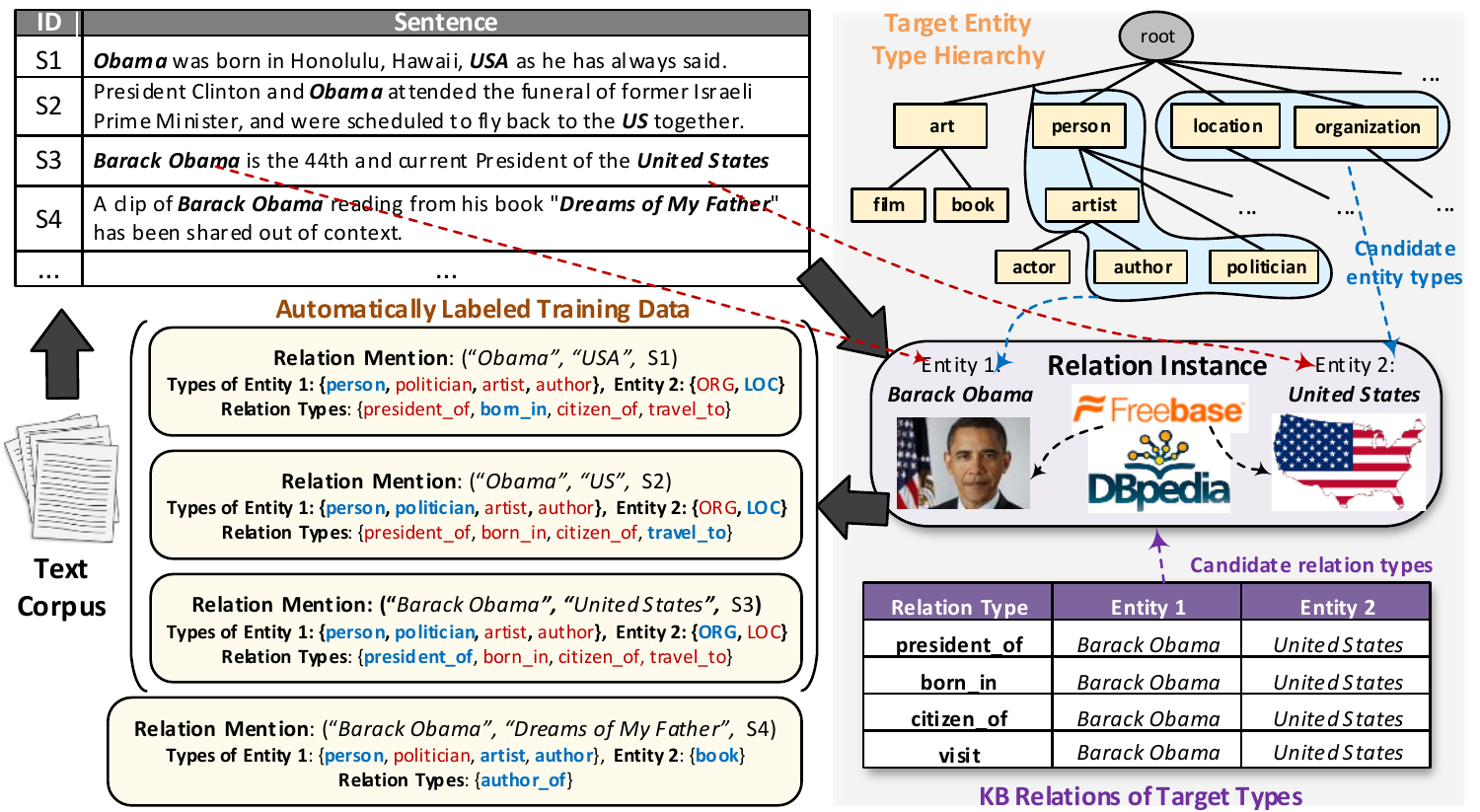}
%\vspace{-0.5cm}
\caption{Current systems find relations (\textit{Barack Obama}, \textit{United States}) mentioned in sentences S1-S3 and assign the same relation types (entity types) to all relation mentions (entity mentions), when only some types are correct for context (highlighted in blue font).}
\label{figure:motivated_example}
\end{small}
\vspace{-0.1cm}
\end{figure}

%%% Distant supervision:
% - motivation:  issues of traditional systems; wide usage & advantage of DS
A major challenge in joint extraction of typed entities and relations is to design \textit{domain-independent} systems that will apply to text corpora from different domains in the \textit{absence of human-annotated, domain data}. The process of manually labeling a training set with a large number of entity and relation types is too expensive and error-prone. The rapid emergence of large, domain-specific text corpora (\eg, news, scientific publications, social media content) calls for methods that can jointly extract entities and relations of target types with minimal or no human supervision. 

% - Related Studies
Towards this goal, there are broadly two kinds of efforts: weak supervision and distant supervision. Weak supervision~\cite{bunescu2007learning,nakashole2013fine,etzioni2004web} relies on a small set of manually-specified seed instances (or patterns) that are applied in bootstrapping learning to identify more instances of each type. This assumes seeds are unambiguous and sufficiently frequent in the corpus, which requires careful seed selection by human~\cite{bach2007review}. Distant supervision~\cite{mintz2009distant,riedel2010modeling,
hoffmann2011multiR,surdeanu2012MIME} generates training data automatically by aligning texts and a knowledge base (KB) (see Fig.~\ref{figure:motivated_example}). The typical workflow is: (1)~detect entity mentions in text; (2)~map detected entity mentions to entities in KB; (3)~assign, to the candidate type set of each entity mention, all KB types of its KB-mapped entity; (4)~assign, to the candidate type set of each entity mention pair, all KB relation types between their KB-mapped entities. The automatically labeled training corpus is then used to infer types of the \textit{remaining} candidate entity mentions and relation mentions (\ie, unlinkable candidate mentions).

%%%%%%%%%%%%%%%%%
\begin{table}[t]
%\vspace{-0.6cm}
\begin{small}
\begin{center}
\hspace*{-0.3cm}
\begin{tabularx}{1.03\linewidth}{l|ccc}
\hline
\begin{scriptsize}
\textbf{Dataset}\end{scriptsize} & \begin{scriptsize}
\textbf{NYT}~\cite{riedel2010modeling}\end{scriptsize}  & \begin{scriptsize}\textbf{Wiki-KBP}~\cite{ellislinguistic},\end{scriptsize} & \begin{scriptsize}\textbf{BioInfer}~\cite{pyysalo2007bioinfer}\end{scriptsize}\\ \hline
\# of entity types & 47 & 126 & 2,200 \\
noisy entity mentions (\%) & 20.32 & 28.31 & 59.80 \\
\hline
\# of relation types & 24 & 19 & 94 \\
noisy relation mentions (\%) & 15.54 & 8.54 & 41.12 \\
\hline
\end{tabularx}
%\vspace{-0.2cm}
\caption{\textbf{A study of type label noise}. \scriptsize (1): \%entity mentions with multiple \textit{sibling entity types} (\eg, \texttt{\scriptsize actor}, \texttt{\scriptsize singer}) in the given entity type hierarchy; (2): \%relation mentions with multiple \textit{relation types}, for the three experiment datasets.}
\label{table:label_noise_stats}
\vspace{-0.3cm}
\end{center}
\end{small}
\end{table}
%%%%%%%%%%%%%%%%%%

%%% - Problem to study
In this paper, we study the problem of \textit{joint extraction of typed entities and relations with distant supervision}. Given a domain-specific corpus and a set of target entity and relation types from a KB, we aim to detect relation mentions (together with their entity arguments) from text, and categorize each in context by target types or Not-Target-Type (\begin{small}\texttt{None}\end{small}), with distant supervision. 
% - Limitations of related studies (with illustrations)
% 1. label noise in DS 
% 2. error propagation in current incremental pipelines
Current distant supervision methods focus on solving the subtasks separately (\eg, extracting typed entities or relations), and encounter the following limitations when handling the joint extraction task.

\noindent$\bullet$ \textbf{\small Domain Restriction:}
They rely on pre-trained named entity recognizers (or noun phrase chunker) to detect entity mentions. These tools are usually designed for a few general types (\eg, \begin{small}\texttt{person}\end{small}, \begin{small}\texttt{location}\end{small}, \begin{small}\texttt{organization}\end{small}) and require additional human labors to work on specific domains (\eg, scientific publications).

\noindent$\bullet$ \textbf{\small Error Propagation:}
In current extraction pipelines, incorrect entity types generated in entity recognition and typing step serve as features in the relation extraction step (\ie, errors are propagated from upstream components to downstream ones). Cross-task dependencies are ignored in most existing methods.

\noindent$\bullet$ \textbf{\small Label Noise:}
In distant supervision, the context-agnostic mapping from relation (entity) mentions to 
KB relations (entities) may bring false positive type labels (\ie, label noise) into the automatically labeled training corpora and results in inaccurate models.

In Fig.~\ref{figure:motivated_example}, for example, all KB relations between entities \textit{Barack Obama} and \textit{United States}  (\eg, \begin{small}\texttt{born\_in}\end{small}, \begin{small}\texttt{president\_of}\end{small}) are assigned to the relation mention in sentence $S_1$ (while only \begin{small}\texttt{born\_in}\end{small} is correct within the  context). Similarly, all KB types for \textit{Barack Obama} (\eg, \begin{small}\texttt{politician}\end{small}, \begin{small}\texttt{artist}\end{small}) are assigned to the mention ``\textit{Obama}" in $S_1$ (while only \begin{small}\texttt{person}\end{small} is true). Label noise becomes an impediment to learn effective type classifiers. The larger the target type set, the more severe the degree of label noise (see Table~\ref{table:label_noise_stats}).

%% Our solution idea
% - joint modeling also helps reducing the level of label noise
We approach the joint extraction task as follows: (1)~Design a domain-agnostic text segmentation algorithm to detect candidate entity mentions with distant supervision and minimal linguistic assumption (\ie, assuming part-of-speech (POS) tagged corpus is given~\cite{hovy2015mining}). 
(2)~Model the mutual constraints between the types of the relation mentions and the types of their entity arguments, to enable feedbacks between the two subtasks. (3)~Model the true type labels in a candidate type set as latent variables and require only the ``\textit{best}" type (progressively estimated as we learn the model) to be relevant to the mention---this is a less limiting requirement compared with existing multi-label classifiers that assume ``\textit{every}" candidate type is relevant to the mention.

%% Our technical meat
% - use text features extracted from local context to model types of a mention
% - partial-label loss to faithfully model noisy type labels from distant supervision; with the help of text features from local context
% - Translation-based objective to model the mutual constraints between relations and their entity arguments; it also helps provide more information to denoise type labels.
%
To integrate these elements of our approach, a novel framework, \textsc{CoType}, is proposed. It first runs POS-constrained text segmentation using positive examples from KB to mine quality entity mentions, and forms candidate relation mentions (Sec.~\ref{subsec:candidate_generation}). Then \textsc{CoType} performs entity linking to map candidate relation (entity) mentions to KB relations (entities) and obtain the KB types. We formulate a global objective to \textit{jointly} model (1) corpus-level co-occurrences between \textit{linkable} relation (entity) mentions and text features extracted from their local contexts; (2) associations between mentions and their KB-mapped type labels; and (3) interactions between relation mentions and their entity arguments. In particular, we design a novel partial-label loss to model the noisy mention-label associations in a robust way, and adopt translation-based objective to capture the entity-relation interactions. Minimizing the objective yields two low-dimensional spaces (for entity and relation mentions, respectively), where, in each space, objects whose types are semantically close also have similar representation (see Sec.~\ref{subsec:model}). With the learned embeddings, we can efficiently estimate the types for the remaining \textit{unlinkable} relation mentions and their entity arguments (see Sec.~\ref{subsec:algorithm}).

\begin{figure*}
\centering
\vspace{-0.3cm}
\includegraphics[width = 174 mm]{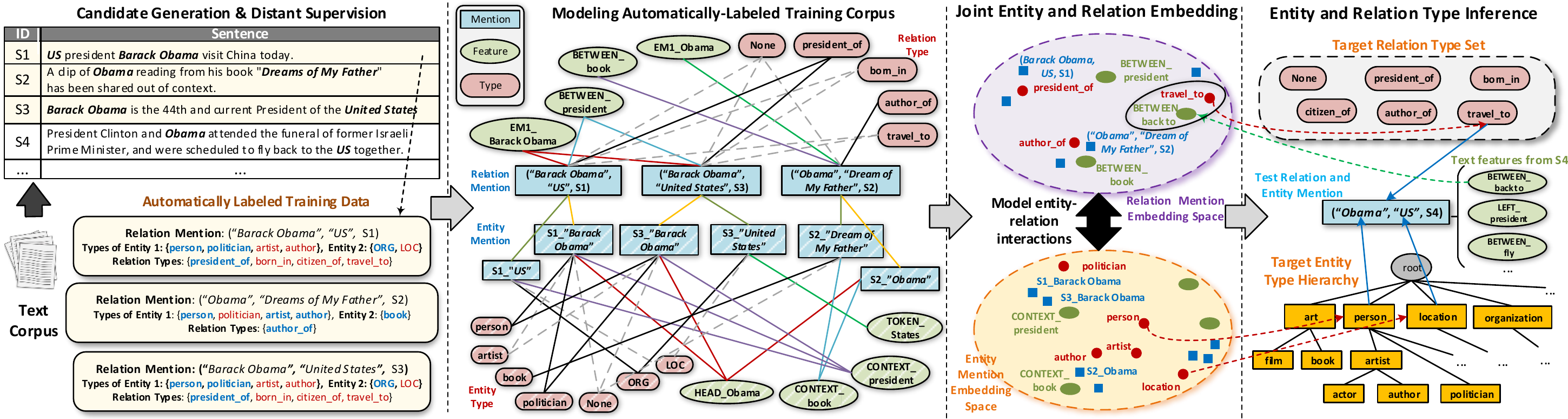}
%\vspace{-0.2cm}
\caption{Framework Overview of \textsc{CoType}.}
\label{figure:framework_overview}
%\vspace{-0.3cm}
\end{figure*}

%%% Contributions
The major contributions of this paper are as follows:
%\vspace{-0.1cm}
\begin{enumerate}[leftmargin=12pt]\itemsep+0.1cm
\item A novel distant-supervision framework, \textsc{CoType}, is proposed to extract typed entities and relations in domain-specific corpora with minimal linguistic assumption. (Fig.~\ref{figure:framework_overview}.)
\item A domain-agnostic text segmentation algorithm is developed to detect entity mentions using distant supervision. (Sec.~\ref{subsec:candidate_generation}) 
\item A joint embedding objective is formulated that models mention-type association, mention-feature co-occurrence, entity-relation cross-constraints in a noise-robust way. (Sec.~\ref{subsec:model}) 
\item Experiments with three public datasets demonstrate that \textsc{CoType} improves the performance of state-of-the-art systems of entity typing and relation extraction significantly, demonstrating robust domain-independence.(Sec.~\ref{sec:experiments})
\end{enumerate}
%\vspace{-0.2cm}

%%%%%%%%%%%%%%%%%%%%%%%
% problem
%%%%%%%%%%%%%%%%%%%%%%%
\section{Background and Problem}
\label{sec:problem}
%\vspace{-0.0cm}

%% input
The input to our proposed \textsc{CoType} framework is a POS-tagged text corpus \begin{small}$\D$\end{small}, a knowledge bases \begin{small}$\Psi$\end{small} (\eg, Freebase~\cite{bollacker2008freebase}), a target entity type hierarchy \begin{small}$\Y$\end{small} and a target relation type set \begin{small}$\R$\end{small}.  The target type set \begin{small}$\Y$\end{small} (set \begin{small}$\R$\end{small}) covers a subset of entity (relation) types that the users are interested in from \begin{small}$\Psi$\end{small}, \ie, \begin{small}$\Y\subset\Y_\Psi$\end{small} and \begin{small}$\R\subset\R_\Psi$\end{small}.

\smallskip\noindent
\textbf{\small \textsf{Entity and Relation Mention.}}
% entity mention, relation instance, relation mention
An \textit{entity mention} (denoted by $m$) is a token span in text which represents an entity $e$. A \textit{relation instance} \begin{small}$r(e_1,e_2, \ldots, e_n)$\end{small} denotes some type of relation \begin{small}$r\in\R$\end{small} between multiple entities. In this work, we focus on binary relations, \ie, $r(e_1,e_2)$.
We define a \textit{relation mention} (denoted by $z$) for some relation instance \begin{small}$r(e_1,e_2)$\end{small} as a (ordered) pair of entities mentions of $e_1$ and $e_2$ in a sentence $s$, and represent a relation mention with entity mentions $m_1$ and $m_2$ in sentence $s$ as \begin{small}$z=(m_1, m_2, s)$\end{small}.

\smallskip\noindent
\textbf{\small \textsf{Knowledge Bases and Target Types.}}
A KB with a set of entities \begin{small}$\E_\Psi$\end{small} contains human-curated facts on both relation instances \begin{small}$\I_\Psi = \{r(e_1,e_2)\} \subset \R_\Psi \times \E_\Psi\times
\E_\Psi$\end{small}, and entity-type facts \begin{small}$\T_\Psi=\{(e,y)\}\subset \E_\Psi\times\Y_\Psi$\end{small}. \textit{Target entity type hierarchy} is a tree where nodes represent entity types of interests from the set \begin{small}$\Y_\Psi$\end{small}. An entity mention may have multiple types, which together constitute one \textit{type-path} (not required to end at a leaf) in the given type hierarchy.
In existing studies, several entity type hierarchies are manually constructed using Freebase~\cite{lee2007fine,gillick2014context} or WordNet~\cite{yosef2012hyena}. \textit{Target relation type set} is a set of relation types of interests from the set \begin{small}$\R_\Psi$\end{small}.

\smallskip\noindent
\textbf{\small \textsf{Automatically Labeled Training Data.}}
Let \begin{small}$\M=\{m_i\}_{i=1}^N$\end{small} denote the set of entity mentions extracted from corpus \begin{small}$\D$\end{small}. Distant supervision maps \begin{small}$\M$\end{small} to KB entities \begin{small}$\E_\Psi$\end{small} with an entity disambiguation system~\cite{mendes2011dbpedia,hoffart2011robust} and heuristically assign type labels to the mapped mentions. In practice, only a small number of entity mentions in set \begin{small}$\M$\end{small} can be mapped to entities in \begin{small}$\E_\Psi$\end{small} (\ie, \textit{linkable entity mentions}, denoted by \begin{small}$\M_L$\end{small}). As reported in~\cite{ren2015clustype,lin2012no}, the ratios of \begin{small}$\M_L$\end{small} over \begin{small}$\M$\end{small} are usually \textit{lower than 50\%} in domain-specific corpora.

Between any two linkable entity mentions $m_1$ and $m_2$ in a sentence, a relation mention $z_i$ is formed if there exists one or more KB relations between their KB-mapped entities $e_1$ and $e_2$. Relations between $e_1$ and $e_2$ in KB are then associated to $z_i$ to form its candidate relation type set \begin{small}$\R_i$\end{small}, \ie, \begin{small}$\R_i=\{r~|~r(e_1, e_2)\in\R_\Psi\}$\end{small}.
In a similar way, types of $e_1$ and $e_2$ in KB are associated with $m_1$ and $m_2$ respectively, to form their candidate entity type sets \begin{small}$\Y_{i,1}$\end{small} and \begin{small}$\Y_{i,2}$\end{small}, where \begin{small}$\Y_{i,x}=\{y~|~(e_x, y)\in\Y_\Psi\}$\end{small}.
Let \begin{small}$\Z_L=\{z_i\}_{i=1}^{N_L}$\end{small} denote the set of extracted relation mentions that can be mapped to KB.
Formally, we represent the automatically labeled training corpus for the joint extraction task, denoted as \begin{small}$\D_L$\end{small}, using a set of tuples \begin{small}$\D_L = \{(z_i, \R_i, \Y_{i,1}, \Y_{i,2})\}_{i=1}^{N_L}$\end{small}.

\smallskip\noindent
\textbf{\small \textsf{Problem Description.}}
% define M_U
By pairing up entity mentions (from set \begin{small}$\M$\end{small}) within each sentence in \begin{small}$\D$\end{small}, we generate a set of \textit{candidate relation mentions}, denoted as \begin{small}$\Z$\end{small}. Set \begin{small}$\Z$\end{small} consists of (1) linkable relation mentions \begin{small}$\Z_L$\end{small}, (2) unlinkable (true) relation mentions, and (3) false relation mention (\ie, no target relation expressed between).

Let \begin{small}$\Z_U$\end{small} denote the set of \textit{unlabeled} relation mentions in (2) and (3) (\ie, \begin{small}$\Z_U = \Z \setminus\Z_L$\end{small}).
% define the joint extraction task
Our main task is to determine the relation type label (from the set \begin{small}$\R\cup\{\texttt{None}\}$\end{small}) for each relation mention in set \begin{small}$\Z_U$\end{small}, and the entity type labels (either a single type-path in \begin{small}$\Y$\end{small} or \begin{small}\texttt{None}\end{small}) for each entity mention argument in \begin{small}$z\in\Z_U$\end{small}, using the automatically labeled corpus \begin{small}$\D_L$\end{small}. Formally, we define the joint extraction of typed entities and relations task as follows.
%%%%
\begin{definition}[Problem Definition]
\textbf{Given} a POS-tagged corpus \begin{small}$\D$\end{small}, a KB \begin{small}$\Psi$\end{small}, a target entity type hierarchy \begin{small}$\Y\subset\Y_\Psi$\end{small} and a target relation type set \begin{small}$\R\subset\R_\Psi$\end{small}, the joint extraction task \textbf{aims to} (1) detect entity mentions \begin{small}$\M$\end{small} from \begin{small}$\D$\end{small}; (2) generate training data \begin{small}$\D_L$\end{small} with KB \begin{small}$\Psi$\end{small}; and (3) estimate a relation type \begin{small}$r^*\in\R\cup$\{\texttt{None}\}\end{small} for each test relation mention \begin{small}$z\in\Z_U$\end{small} and a single type-path \begin{small}$\Y^*\subset\Y$\end{small} (or \texttt{None}) for each entity mention in $z$, using \begin{small}$\D_L$\end{small} and its context $s$.
\end{definition}

\smallskip\noindent
\textbf{\small \textsf{Non-goals.}}
% entity disambiguation is out of scope
%Label noise may come from wrong mapping of mentions to KB entities.
This work relies on an entity linking system~\cite{mendes2011dbpedia} to provide disambiguation function, but we do not address their limits here (\eg, label noise introduced by wrongly mapped KB entities). We also assume human-curated target type hierarchies are given (It is out of the scope of this study to generate the type hierarchy).

%%%%%%%%%%%%%%%%%%%%%%%
% method
%%%%%%%%%%%%%%%%%%%%%%%
\section{The CoType Framework}
\label{sec:method}
%% intuitive ideas of the proposed framework
This section lays out the proposed framework.
% challenges of current task: 1. label noise in R_i, Y_i,1, Y_i,2 of z_i in Z_L; 2. dependencies between z_i and m_i,1, m_i,2 when modeling R_i, Y_i,1, Y_i,2.
% + simple ideas (and why not effective)
The joint extraction task poses two unique challenges. First, type association in distant supervision between linkable entity (relation) mentions and their KB-mapped entities (relations) is \textit{context-agnostic}---the candidate type sets \begin{small}$\{\R_i, \Y_{i,1}, \Y_{i,2}\}$\end{small} contain ``false" types. Supervised learning~\cite{guodong2005exploring,gormley2015improved} may generate models biased to the incorrect type labels~\cite{ren2016label}.
Second, there exists dependencies between relation mentions and their entity arguments (\eg, type correlation). Current systems  formulates the task as a \textit{cascading} supervised learning problem and may suffer from error propagation.
%---entity types are first estimated then serve as features in relation classification---

% High-level priciples + technical ideas of CoType
Our solution casts the type prediction task as \textit{weakly-supervised learning} (to model the relatedness between mentions and their candidate types \textit{in contexts}) and uses \textit{relational learning} to capture interactions between relation mentions and their entity mention argument \textit{jointly}, based on the redundant text signals in a large corpus.

Specifically, \textsc{CoType} leverages partial-label learning~\cite{nguyen2008classification} to faithfully model mention-type association using text features extracted from mentions' local contexts. It uses the translation embedding-based objective~\cite{bordes2013translating} to model the mutual type dependencies between relation mentions and their entity (mention) arguments.

\medskip
\noindent \textsf{\textbf{Framework Overview.}}
We propose a \textit{embedding-based} framework with distant supervision (see also Fig.~\ref{figure:framework_overview}) as follows:

\begin{enumerate}[leftmargin=12pt]\itemsep+0.1cm
%\vspace{-0.15cm}
\item Run POS-constrained text segmentation algorithm on POS-tagged corpus \begin{small}$\D$\end{small} using positive examples obtained from KB, to detect candidate entity mentions \begin{small}$\M$\end{small} (Sec.~\ref{subsec:candidate_generation}). 

\item Generate candidate relation mentions \begin{small}$\Z$\end{small} from \begin{small}$\M$\end{small}, extract text features for each relation mention \begin{small}$z\in\Z$\end{small} and their entity mention argument (Sec.~\ref{subsec:candidate_generation}). Apply distant supervision to generate labeled training data \begin{small}$\D_L$\end{small} (Sec.~\ref{sec:problem}).

\item Jointly embed relation and entity mentions, text features, and type labels into two low-dimensional spaces (for entities and relations, respectively) where, in each space, close objects tend to share the same types (Sec.~\ref{subsec:model}).

\item Estimate type labels \begin{small}$r^*$\end{small} for each test relation mention \begin{small}$z\in\Z_U$\end{small} and type-path \begin{small}$\Y^*$\end{small} for each test entity mention \begin{small}$m$ in $\Z_U$\end{small} from learned embeddings, by searching the target type set \begin{small}$\Y$\end{small} or the target type hierarchy \begin{small}$\R$\end{small} (Sec.~\ref{subsec:algorithm}).
\end{enumerate}

\subsection{Candidate Generation}
\label{subsec:candidate_generation}
% 0.4 page
%This section introduces the candidate generation process for detecting entity and relation mentions from POS-tagged corpus \begin{small}$\D$\end{small}, and the generation of labeled corpus \begin{small}$\D_L$\end{small} as well as text features for the candidate mentions.

%\smallskip
\noindent
\textsf{\small\textbf{Entity Mention Detection.}}
%% POS-guided text segementation for candidate entity mentions
% - remember to mention why pre-trained NER is not compared here.
Traditional entity recognition systems\\~\cite{finkel2005incorporating,nadeau2007survey} rely on a set of linguistic features (\eg, dependency parse structures of a sentence) to train sequence labeling models (for a few common entity types). However, sequence labeling models trained on automatically labeled corpus \begin{small}$\D_L$\end{small} may not be effective, as distant supervision only annotates a small number of entity mentions in \begin{small}$\D_L$\end{small} (thus generates a lot of ``false negative" token tags). To address domain restriction, we develop a distantly-supervised text segmentation algorithm for \textit{domain-agnostic entity detection}. By using quality examples from KB as guidance, it partitions sentences into segments of entity mentions and words, by incorporating (1) corpus-level concordance statistics; (2) sentence-level lexical signals; and (3) grammatical constraints (\ie, POS tag patterns). 

We extend the methdology used in~\cite{liu2015mining,el2014scalable} to model the \textit{segment quality}  (\ie, ``how likely a candidate segment is an entity mention") as a combination of \textit{phrase quality} and \textit{POS pattern quality}, and use positive examples in \begin{small}$\D_L$\end{small} to estimate the segment quality.
The workflow is as follows: (1) mine frequent contiguous patterns for both word sequence and POS tag sequence up to a fixed length from POS-tagged corpus \begin{small}$\D$\end{small}; (2) extract features including corpus-level concordance and sentence-level lexical signals to train two random forest classifiers~\cite{liu2015mining}, for estimating quality of candidate phrase and candidate POS pattern; (3) find the best segmentation of \begin{small}$\D$\end{small} using the estimated segment quality scores (see Eq.~\eqref{eq:segmentation}); and (4) compute rectified features using the segmented corpus and repeat steps (2)-(4) until the result converges.
\begin{equation}
\label{eq:segmentation}
p\big(b_{t+1}, c~|~b_{t}\big)=p\big(b_{t+1} - b_{t}\big)\cdot p\big(c~|~b_{t+1} - b_{t}\big)\cdot Q(c)
\end{equation}

Specifically, we find the best segmentation \begin{small}$S_d$\end{small} for each document \begin{small}$d$\end{small} (in \begin{small}$\D$\end{small}) by maximizing the ``joint segmentation quality", defined as \begin{scriptsize}$\sum_{d}^\D\log~p(S_d, d)=\sum_{d}^\D\sum_{t=1}^{|d|}\log~p\big(b_{t+1}^{(d)}, c^{(d)}~|~b_{t}^{(d)}\big)$\end{scriptsize}, where \begin{scriptsize}$p\big(b_{t+1}^{(d)}, c^{(d)}~|~b_{t}^{(d)}\big)$\end{scriptsize} denote the probability that segment \begin{small}$c^{(c)}$\end{small} (with starting index \begin{small}$b_{t+1}^{(d)}$\end{small} and ending index in document \begin{small}$d$\end{small}) is a good entity mention, as defined in Eq.~\eqref{eq:segmentation}. The first term in Eq.~\eqref{eq:segmentation} is a segment length prior, the second term measures how likely segment \begin{small}$c$\end{small} is generated given a length \begin{small}$(b_{t+1}-b_{t})$\end{small} (to be estimated), and the third term denotes the segment quality. In this work, we define function \begin{small}$Q(c)$\end{small} as the \textit{equally weighted combination of the phrase quality score and POS pattern quality score} for candidate segment \begin{small}$c$\end{small}, which is estimated in step (2). The joint probability can be efficiently maximize using Viterbi Training with time complexity linear to the corpus size~\cite{liu2015mining}. The segmentation result provides us a set of candidate entity mentions, forming the set \begin{small}$\M$\end{small}.

\begin{figure}
\centering
\vspace{-0.3cm}
\includegraphics[width = 72 mm]{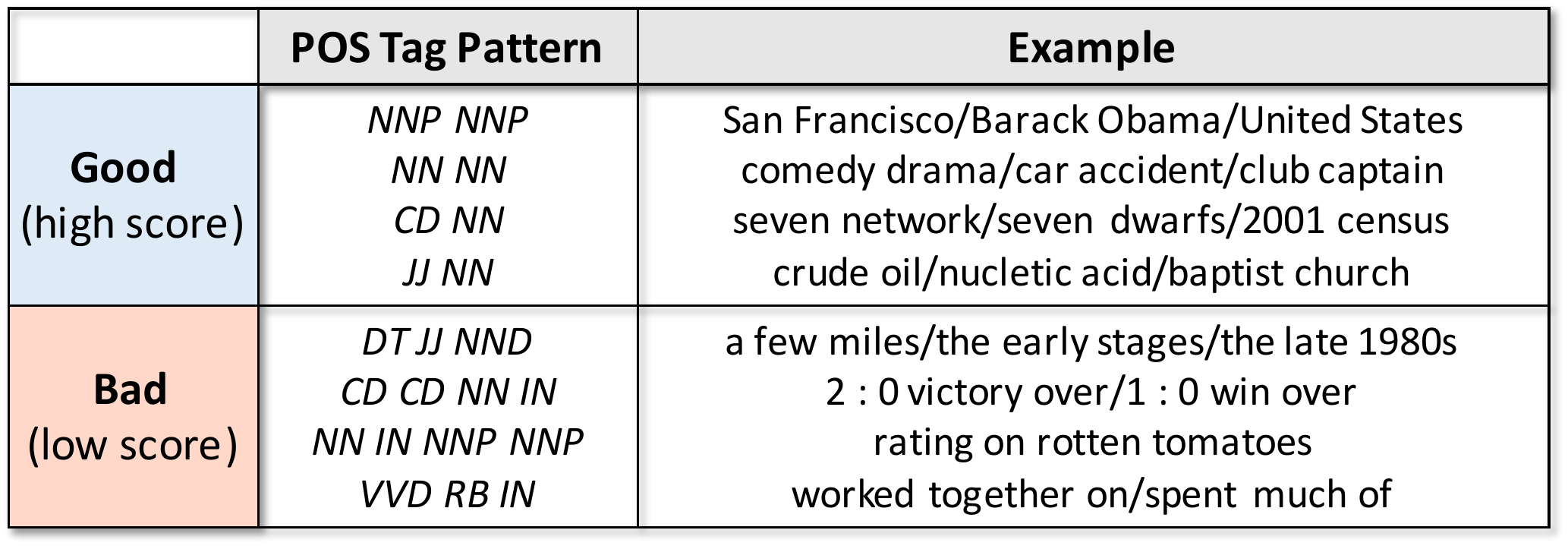}
%\vspace{-0.3cm}
\caption{Example POS tag patterns learned using KB examples.}
\label{figure:pos_pattern}
%\vspace{-0.1cm}
\end{figure}

Table~\ref{table:entity_detection} compares our entity detection module with a sequence labeling model~\cite{ling2012fine} (linear-chain CRF) trained on the labeled corpus \begin{small}$\D_L$\end{small} in terms of F1 score. Fig.~\ref{figure:pos_pattern} show the high/low quality POS patterns learned using entity names found  in \begin{small}$\D_L$\end{small} as examples.

\smallskip\noindent
\textsf{\small\textbf{Relation Mention Generation.}}
%% generate candidate relation mentions
We follow the procedure introduced in Sec.~\ref{sec:problem} to generate the set of candidate relation mentions \begin{small}$\Z$\end{small} from the detected candidate entity mentions \begin{small}$\M$\end{small}: for each pair of entity mentions \begin{small}$(m_a, m_b)$\end{small} found in sentence \begin{small}$s$\end{small}, we form two candidate relation mentions \begin{small}$z_1=(m_a,m_b,s)$\end{small} and \begin{small}$z_2=(m_b,m_a,s)$\end{small}.
Distant supervision is then applied on \begin{small}$\Z$\end{small} to generate the set of KB-mapped relation mentions \begin{small}$\Z_L$\end{small}. Similar to ~\cite{mintz2009distant,hoffmann2011multiR}, we sample 30\% unlinkable relation mentions between two KB-mapped entity mentions (from set \begin{small}$\M_L$\end{small}) in a sentence as examples for modeling \texttt{None} relation label, and sample 30\% unlinkable entity mentions (from set \begin{small}$\M\setminus\M_L$\end{small}) to model \texttt{None} entity label. These negative examples, together with type labels for mentions in \begin{small}$\Z_L$\end{small}, form the automatically labeled data \begin{small}$\D_L$\end{small} for the task.

% how to build Z_L
% - talk about how to add RMs for "None" label; Explain that this way is less likely to include "false negative" examples.

%%%%%%%%%%%%%%%%%
\begin{table}[t]
%\vspace{-0.0cm}
\begin{scriptsize}
\begin{center}
\begin{tabularx}{0.8\linewidth}{l|ccc}
\hline
\begin{scriptsize}
\textbf{Dataset}\end{scriptsize} & \begin{scriptsize}
\textbf{NYT}\end{scriptsize}  & \begin{scriptsize}\textbf{Wiki-KBP}\end{scriptsize} & \begin{scriptsize}\textbf{BioInfer}\end{scriptsize}\\ \hline
%NP Chunker & 0.402 & 0.457 & 0.264 \\
FIGER segmenter~\cite{ling2012fine}$\quad$ & 0.751 & 0.814 & 0.652 \\
Our Approach & \textbf{0.837} & \textbf{0.833} & \textbf{0.785} \\
\hline
\end{tabularx}
%\vspace{-0.2cm}
\caption{Comparison of F1 scores on entity mention detection.}
\label{table:entity_detection}
%\vspace{-0.6cm}
\end{center}
\end{scriptsize}
\end{table}
%%%%%%%%%%%%%%%%%%

\smallskip
\noindent 
\textsf{\small\textbf{Text Feature Extraction.}}
% RM and EM features
To capture the shallow syntax and distributional semantics
of a relation (or entity) mention, we extract various lexical features from both mention itself (\eg, head token) and its context $s$ (\eg, bigram), in the POS-tagged corpus. Table~\ref{table:features} lists the set of text features for relation mention, which is similar to those used in~\cite{mintz2009distant,chan2010exploiting} (excluding the dependency parse-based features and entity type features). We use the same set of features for entity mentions as those used in~\cite{ren2016label,ling2012fine}.
We denote the set of \begin{small}$M_z$\end{small} (\begin{small}$M_m$\end{small}) unique features extracted of relation mentions \begin{small}$\Z_L$\end{small} (entity mentions in \begin{small}$\Z_L$\end{small}) as \begin{small}$\F_{z}=\{f_j\}_{j=1}^{M_z}$\end{small} \big(and \begin{small}$\F_m=\{f_j\}_{j=1}^{M_m}$\end{small}\big).

%%%%%%%%%%%%%%%%%
\begin{table}[t]
\vspace{-0.3cm}
\begin{center}
\hspace*{-0.6cm}
\begin{tiny}
\begin{tabularx}{1.1\linewidth}{lll}
%{>{\hsize=0.5\hsize}X|>{\hsize=1.5\hsize}X| X}
\hline
\textbf{Feature} & \textbf{Description} & \textbf{Example} \\
\hline
Entity mention (EM) head & Syntactic head token of each entity mention & ``\textit{HEAD\_EM1\_Obama}"%, ``\textit{HEAD\_EM2\_States}"
\\ %\hline
Entity Mention Token & Tokens in each entity mention & ``\textit{TKN\_EM1\_Barack}"%, ``\textit{TKN\_EM1\_Obama}"
%, ``TKN\_EM2\_United", ``TKN\_EM2\_States"
\\ %\hline
Tokens between two EMs & Each token between two EMs & ``\textit{was}", ``\textit{elected}", ``\textit{President}", ``\textit{of}", ``\textit{the}"
\\
Part-of-speech (POS) tag & POS tags of tokens between two EMs & ``\textit{VBD}", ``\textit{VBN}", ``\textit{NNP}", ``\textit{IN}", ``\textit{DT}" \\ %\hline
Collocations & Bigrams in left/right 3-word window of each EM & ``\textit{Honolulu native}", ``\textit{native Barack}", ... 
\\ %\hline
Entity mention order & Whether EM 1 is before EM 2 & ``\textit{EM1\_BEFORE\_EM2}"\\ %\hline
Entity mention distance & Number of tokens between the two EMs & ``\textit{EM\_DISTANCE\_5}" \\ %\hline
%\#Entity mentions in between~~~~~ & Number of other EMs in between the two EMs & ``0" \\ %\hline
Entity mention context & Unigrams before and after each EM & ``\textit{native}", ``\textit{was}", ``\textit{the}", ``\textit{in}" \\ %\hline
Special pattern & Occurrence of pattern ``em1\_in\_em2" & ``\textit{PATTERN\_NULL}" \\ %\hline
%Entity type & type of each entity mention & ``EM1\_/person/politician", ``EM2\_/location/country" \\ %\hline
Brown cluster (learned on $\D$) & Brown cluster ID for each token  & ``\textit{8\_1101111}",~``\textit{12\_111011111111}" \\ %\hline
\hline
\end{tabularx}
\end{tiny}
%\vspace{-0.2cm}
\caption{\scriptsize Text features for relation mentions used in this work~\cite{guodong2005exploring,riedel2010modeling} (excluding dependency parse-based features and entity type features). (``\textit{Barack Obama}'', ``\textit{United States}'') is used as an example relation mention from the sentence ``\textit{Honolulu native \textbf{Barack Obama} was elected President of the \textbf{United States} on March 20 in 2008.}".}
\label{table:features}
%\vspace{-0.6cm}
\end{center}
\end{table}
%%%%%%%%%%%%%%%%%

\subsection{Joint Entity and Relation Embedding}
\label{subsec:model}
% 1.5-2 pages (only text)
This section formulates a joint optimization problem for embedding different kinds of interactions between linkable relation mentions \begin{small}$\Z_L$\end{small}, linkable entity mentions \begin{small}$\M_L$\end{small}, entity and relation type labels \{\begin{small}$\R,~\Y$\end{small}\} and text features \{\begin{small}$\F_z,~\F_m$\end{small}\} into a $d$-dimensional \textit{relation vector space} and a $d$-dimensional \textit{entity vector space}. In each space, objects whose types are close to each other should have similar representation (\eg, see the 3rd col. in Fig.~\ref{figure:framework_overview}).

%% intuition technical ideas
As the extracted objects and the interactions between them form a heterogeneous graph (see the 2nd col. in Fig.~\ref{figure:framework_overview}), a simple solution is to embed the whole graph into a \textit{single} low-dimensional space~\cite{he2004locality,ren2015clustype}. However, such a solution encounters several problems: (1) False types in candidate type sets (\ie, false mention-type links in the graph) negatively impact the ability of the model to determine mention's true types; and (2) a single embedding space cannot capture the differences in entity and relation types (\ie, strong link between a relation mention and its entity mention argument does not imply that they have similar types).

In our solution, we propose a novel global objective, which extends a margin-based rank loss~\cite{nguyen2008classification} to model \textit{noisy mention-type associations} and leverages the second-order proximity idea~\cite{tang2015line} to model corpus-level \textit{mention-feature co-occurrences}. In particular, to capture the \textit{entity-relation interactions}, we adopt a translation-based embedding loss~\cite{bordes2013translating} to bridge the vector spaces of entity mentions and relation mentions.

\nop{
%%%%%%%%%%%%%%%%%
\begin{table}[t]
%\vspace{-0.2cm}
\begin{scriptsize}
\begin{center}
\begin{tabularx}{\linewidth}{l|X}
\hline
$\D$ & Automatically generated training corpus \\ %\hline
$\M=\{m_i\}_{i=1}^N$ & Entity mentions in $\D$ (size $N$) \\ %\hline
%$\F_i$ & Text features associated with $m_i$\\ %\hline
%$\M_j$ & Entity mentions associated with $f_j$\\ %\hline
$\Y=\{y_k\}_{k=1}^K$ & Target entity types (size $K$)\\ %\hline
$\Y_i$ & Candidate types of $m_i$\\%\hline
$\Yn_i$ & Non-candidate types of $m_i$, \ie, $\Yn_i=\Y~\backslash~\Y_i$ \\%\hline
$\F=\{f_j\}_{j=1}^M$ & Text features in $\D$ (size $M$)\\ %\hline
$\bu_i \in\R^d$ & Embedding of mention $m_i$ (dim. $d$)\\ %\hline
$\bc_j\in\R^d$ & Embedding of feature $f_j$ (dim. $d$)\\ %\hline
$\bv_k, \bv'_k\in\R^d$ & Embeddings of type $y_k$ on two views (dim. $d$)\\
\hline
\end{tabularx}
%\vspace{-0.3cm}
\caption{Notations.}
\label{table:notation}
%\vspace{-0.2cm}
\end{center}
\end{scriptsize}
\end{table}
%%%%%%%%%%%%%%%%%
}
\medskip
\noindent
\textsf{\small\textbf{Modeling Types of Relation Mentions.}}
%%% 0.4 page
% !!! candidate relation type can be all false positive -- it voilate the partial-label loss assumption so need explanation
We consider both \textit{mention-feature co-occurrences} and \textit{mention-type associations} in the modeling of relation types for relation mentions in set \begin{small}$Z_L$\end{small}.

%%%%%%%%
Intuitively, two relation mentions sharing many text features (\ie, with similar distribution over the set of text features \begin{small}$\F_m$\end{small}) likely have similar relation types;  and text features co-occurring with many relation mentions in the corpus tend to represent close type semantics. We propose the following hypothesis to guide our modeling of corpus-level mention-feature co-occurrences.
%\vspace{-0.1cm}
\begin{hypothesis}[Mention-Feature Co-occurrence]
\label{hypo:co_occurrence}
$\newline$Two entity mentions tend to share similar types (close to each other in the embedding space) if they share many text features in the corpus, and the converse way also holds.
\end{hypothesis}

For example, in column 2 of Fig.~\ref{figure:framework_overview}, (``\textit{Barack Obama}", ``\textit{US}", $S_1$) and (``\textit{Barack Obama}", ``\textit{United States}", $S_3$) share multiple features including context word ``\textit{president}" and first entity mention argument ``\textit{Barack Obama}", and thus they are likely of the same relation type (\ie, \begin{small}\texttt{president\_of}\end{small}). 

Formally, let vectors \begin{small}$\bz_i,~\bc_j\in\RR^d$\end{small} represent relation mention \begin{small}$z_i\in\Z_L$\end{small} and text feature \begin{small}$f_j\in\F_z$\end{small} in the $d$-dimensional \textit{relation embedding space}. Similar to the distributional hypothesis~\cite{mikolov2013distributed} in text corpora, we apply second-order proximity~\cite{tang2015line} to model the idea that \textit{objects with similar distribution over neighbors are similar to each other} as follows.
%\vspace{-0.15cm}
\begin{align}
\label{eq:mention_feature_obj}
\L_{ZF} = -\sum_{z_i\in\Z_L}\sum_{f_j\in\F_z}w_{ij}\cdot\log~p(f_j|z_i),
\end{align}
where \begin{small}$p(f_j|z_i) = \e(\bz_i^T\bc_j)\big/\sum_{f^{\prime}\in\F_z}\e(\bz_{i}^T\bc_{j'})$\end{small} denotes the probability of \begin{small}$f_j$\end{small} generated by \begin{small}$z_i$\end{small}, and \begin{small}$w_{ij}$\end{small} is the co-occurrence frequency between \begin{small}$(z_i, f_j)$\end{small} in corpus \begin{small}$\D$\end{small}. Function \begin{small}$\L_{ZF}$\end{small} in Eq.~\eqref{eq:mention_feature_obj} enforces the conditional probability specified by embeddings, \ie, \begin{small}$p(\cdot|z_i)$\end{small} to be close to the empirical distribution.

To perform efficient optimization by avoiding summation over all features, we adopt negative sampling strategy~\cite{mikolov2013distributed} to sample multiple \textit{false} features for each \begin{small}$(z_i,f_j)$\end{small}, according to some \textit{noise distribution} \begin{small}$P_n(f)\propto D_f^{3/4}$\end{small}~\cite{mikolov2013distributed} (with \begin{small}$D_f$\end{small} denotes the number of relation mentions co-occurring with $f$). Term \begin{small}$\log~p(f_j|z_i)$\end{small} in Eq.~\eqref{eq:mention_feature_obj} is replaced with the term as follows.
%\vspace{-0.2cm}
\begin{align}
\label{eq:mention_feature_obj_neg}
\log~\sigma(\bz_i^T\bc_j) + \sum_{v=1}^V\EE_{f_{j'}\sim P_n(f)}\big[\log~\sigma(-\bz_i^T\bc_{j'})\big],
\end{align}
where \begin{small}$\sigma(x)=1/\big(1+\exp(-x)\big)$\end{small} is the sigmoid function. The first term in Eq.~\eqref{eq:mention_feature_obj_neg} models the observed co-occurrence, and the second term models the \begin{small}$Z$\end{small} negative feature samples.

%%%%%%%
In \begin{small}$D_L$\end{small}, each relation mention \begin{small}$z_i$\end{small} is heuristically associated with a set of candidate types \begin{small}$\R_i$\end{small}. 
Existing embedding methods rely on either the \textit{local consistent assumption}~\cite{he2004locality} (\ie, objects strongly connected tend to be similar) or the \textit{distributional assumption}~\cite{mikolov2013distributed} (\ie, objects sharing similar neighbors tend to be similar) to model object associations. However, some associations between \begin{small}$z_i$\end{small} and \begin{small}$r\in\R_i$\end{small} are ``false" associations and adopting the above assumptions may incorrectly yield mentions of different types having similar vector representations. For example, in Fig.~\ref{figure:motivated_example}, mentions (``\textit{Obama}'', ``\textit{USA}'', $S_1$)  and (``\textit{Obama}'', ``\textit{US}'', $S_2$) have several candidate types in common (thus high distributional similarity), but their true types are different (\ie, \begin{small}\texttt{born\_in}\end{small} vs. \texttt{\small travel\_to}).
% explanation + example illustration

We specify the likelihood of ``\textit{whether the association between a relation mention and its candidate entity type being true}" as the \textit{relevance} between these two kinds of objects (measured by the similarity between their current estimated embedding vectors). To impose such idea, we model the associations between each linkable relation mention \begin{small}$z_i$\end{small} (in set \begin{small}$\Z_L$\end{small}) and its noisy candidate relation type set \begin{small}$\R_i$\end{small} based on the following hypothesis.
%\vspace{-0.05cm}
\begin{hypothesis}[Partial-Label Association]
\label{hypo:partial_label}
$\newline$
A relation mention's embedding vector should be more similar (closer in the low-dimensional space) to its ``most relevant" candidate type, than to any other non-candidate type.
\end{hypothesis}

% technical details to enforce the hypothesis
Specifically, we use vector \begin{small}$\br_k\in\RR^d$\end{small} to represent relation type \begin{small}$r_k\in\R$\end{small} in the embedding space. 
The similarity between \begin{small}$(z_i,r_k)$\end{small} is defined as the dot product of their embedding vectors, \ie, \begin{small}$\phi(z_i,r_k)=\bz_i^T\br_k$\end{small}. We extend the margin-based loss in~\cite{nguyen2008classification} and define a partial-label loss \begin{small}$\ell_i$\end{small} for each relation mention \begin{small}$z_i\in\M_L$\end{small} as follows.
%\vspace{-0.1cm}
\begin{align}
\label{eq:partial_label_loss}
\ell_{i} = \max\Big\{0, 1 - \Big[\max_{r\in\R_i}\phi(z_i, r)  - \max_{r'\in \Rn_i}\phi(z_i, r')\Big]\Big\}.
\end{align}

% intuition & example
The intuition behind Eq.~\eqref{eq:partial_label_loss} is that: for relation mention \begin{small}$z_i$\end{small}, the maximum similarity score associated with its candidate type set \begin{small}$\R_i$\end{small} should be greater than the maximum similarity score associated with any other \textit{non-candidate types} \begin{small}$\Rn_i=\R\setminus\R_i$\end{small}. Minimizing \begin{small}$\ell_i$\end{small} forces \begin{small}$z_i$\end{small} to be embedded closer to the \textit{most} ``\textit{relevant}" type in \begin{small}$\R_i$\end{small}, than to any other non-candidate types in \begin{small}$\Rn_i$\end{small}. This contrasts sharply with multi-label learning~\cite{ling2012fine}, where \begin{small}$m_i$\end{small} is embedded closer to \textit{every} candidate type than any other non-candidate type.

To faithfully model the types of relation mentions, we integrate the modeling of mention-feature co-occurrences and mention-type associations by the following objective.
%\vspace{-0.15cm}
\begin{align}
\label{eq:relation_mention_obj}
O_{Z} = \L_{ZF} + \sum_{i=1}^{N_L} \ell_i + \frac{\lambda}{2} \sum_{i=1}^{N_L}\|\bz_i\|_2^2 + \frac{\lambda}{2} \sum_{k=1}^{K_r} \|\br_k\|_2^2,
\end{align}
where tuning parameter \begin{small}$\lambda > 0$\end{small} on the regularization terms is used to control the scale of the embedding vectors.

\begin{figure}
\centering
%\vspace{-0.7cm}
\includegraphics[width = 78 mm]{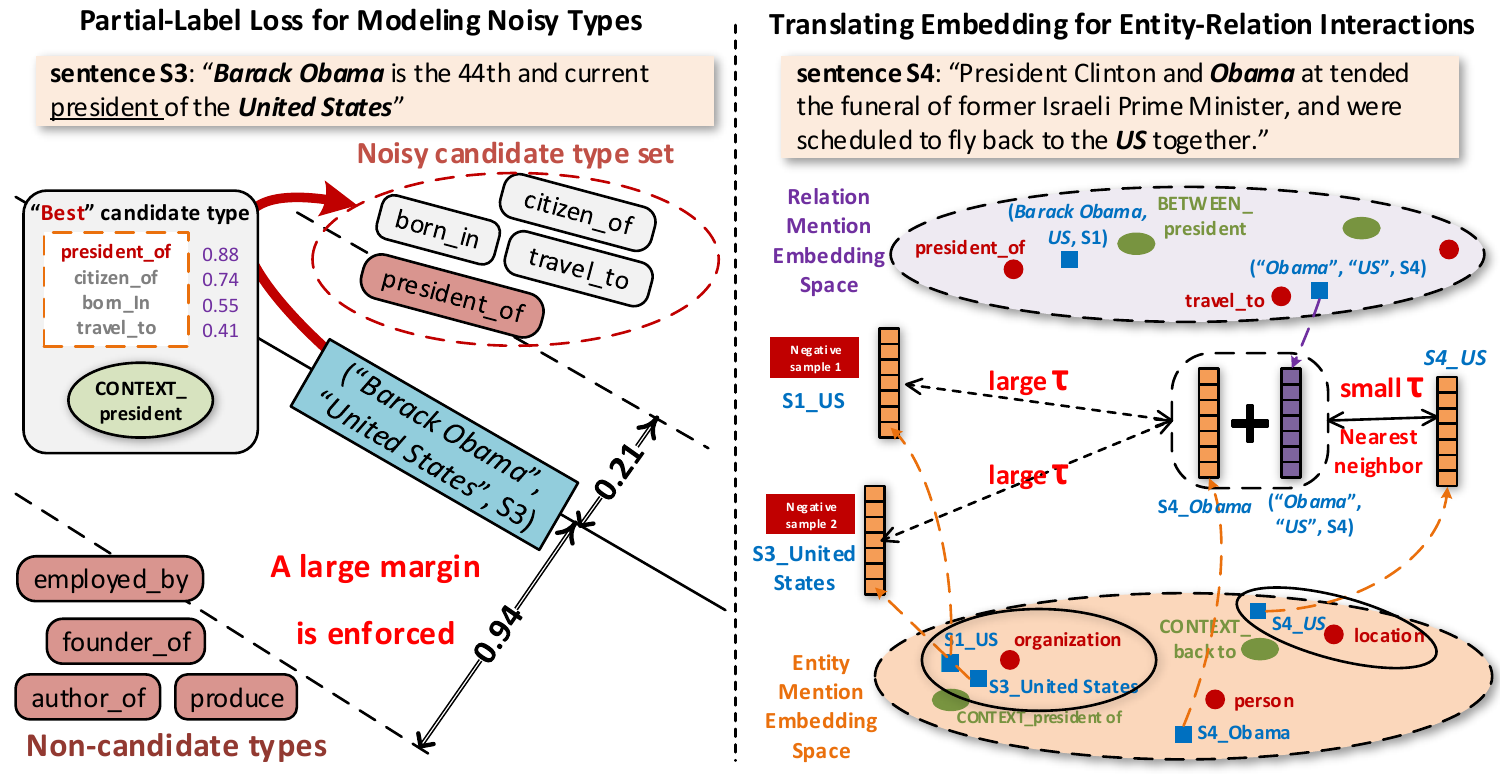}
%\vspace{-0.2cm}
\caption{Illustrations of the partial-label associations, Hypothesis~\ref{hypo:partial_label} (the left col.), and the entity-relation interactions, Hypothesis~\ref{hypo:entity_relation} (the right col.).}
\label{figure:tech_idea_illustration}
%\vspace{-0.4cm}
\end{figure}

By doing so, text features, as complements to mention's candidate types, also participate in modeling the relation mention embeddings, and help identify a mention's most relevant type---mention-type relevance is progressively estimated during model learning. For example, in the left column of Fig.~\ref{figure:tech_idea_illustration}, context words ``\textit{president}"helps infer that relation type \begin{small}\texttt{president\_of}\end{small} is more relevant (\ie, higher similarity between the embedding vectors) to relation mention (``\textit{Mr. Obama}", ``\textit{USA}", $S_2$), than type \begin{small}\texttt{born\_in}\end{small} does.

%%%%%%%
\medskip
\noindent
\textsf{\small\textbf{Modeling Types of Entity Mentions.}}
% 0.2 page
In a way similar to the modeling of types for relation mentions, we follow Hypotheses~\ref{hypo:co_occurrence} and \ref{hypo:partial_label} to model types of entity mentions. 
In Fig.~\ref{figure:framework_overview} (col. 2),  for example, entity mentions ``\textit{$S_1$\_Barack Obama}" and ``\textit{$S_3$\_Barack Obama}" share multiple text features in the corpus, including head token ``\textit{Obama}" and context word ``\textit{president}", and thus tend to share the same entity types like \begin{small}\texttt{politician}\end{small} and \begin{small}\texttt{person}\end{small} (\ie, Hypothesis~\ref{hypo:co_occurrence}).
Meanwhile, entity mentions ``\textit{$S_1$\_Barack Obama}" and ``\textit{$S_2$\_Obama}" have the same candidate entity types but share very few text features in common. This implies that likely their true type labels are different. Relevance between entity mentions and their true type labels should be progressively estimated based on the text features extracted from their local contexts (\ie, Hypothesis~\ref{hypo:partial_label}).

Formally, let vectors \begin{small}$\bm_i,\bc'_j,\by_k\in\RR^d$\end{small} represent entity mention \begin{small}$m_i\in\M_L$\end{small}, text features (for entity mentions) \begin{small}$f_j\in\F_m$\end{small}, and entity type \begin{small}$y_k\in\Y$\end{small} in a $d$-dimensional \textit{entity embedding space}, respectively. We model the corpus-level co-occurrences between entity mentions and text features by second-order proximity as follows.
%\vspace{-0.1cm}
\begin{align}
\label{eq:entity_mention_feature_obj_neg}
\L_{MF} = -\sum_{m_i\in\M_L}\sum_{f_j\in\F_m}w_{ij}\cdot\log~p(f_j|m_i),
\end{align}
where the conditional probability term \begin{small}$\log~p(f_j|m_i)$\end{small} is defined as \begin{small}$\log~p(f_j|m_i) = \log~\sigma(\bm_i^T\bc'_j) + \sum_{v=1}^V\EE_{f_{j'}\sim P_n(f)}\big[\log~\sigma(-\bm_i^T\bc'_{j'})\big]$\end{small}.
By integrating the term \begin{small}$\L_{MF}$\end{small} with partial-label loss \begin{small}$\ell'_{i} = \max\big\{0, 1 - \big[\max_{y\in\Y_i}\phi(m_i, y)  - \max_{y'\in \Yn_i}\phi(m_i, y')\big]\big\}
$\end{small} for \begin{small}$N'_L$\end{small} unique linkable entity mentions (in set \begin{small}$\M_L$\end{small}), we define the objective function for modeling types of entity mentions as follows.
%\vspace{-0.15cm}
\begin{align}
\label{eq:entity_mention_obj}
O_{M} = \L_{MF} + \sum_{i=1}^{N'_L} \ell'_i + \frac{\lambda}{2} \sum_{i=1}^{N'_L}\|\bm_i\|_2^2 + \frac{\lambda}{2} \sum_{k=1}^{K_y} \|\by_k\|_2^2.
\end{align}
Minimizing the objective \begin{small}$O_M$\end{small} yields an entity embedding space where, in that space, objects (\eg, entity mentions, text features) close to each other will have similar types.

\medskip\noindent 
\textsf{\small\textbf{Modeling Entity-Relation Interactions.}}
% 0.3 page
% explanation + example illustration
In reality, there exists different kinds of interactions between a relation mention \begin{small}$z=(m_1,m_2,s)$\end{small} and its entity mention arguments \begin{small}$m_1$\end{small} and \begin{small}$m_2$\end{small}. One major kind of interactions is the correlation between relation and entity types of these objects---entity types of the two entity mentions provide good hints for determining the relation type of the relation mention, and vice versa.
For example, in Fig.~\ref{figure:tech_idea_illustration} (right column), knowing that entity mention ``\textit{$S_4$\_US}" is of type \begin{small}\texttt{location}\end{small} (instead of \begin{small}\texttt{organization}\end{small}) helps determine that relation mention (``\textit{Obama}", ``\textit{US}", $S_4$) is more likely of relation type \begin{small}\texttt{travel\_to}\end{small}, rather than relation types like \begin{small}\texttt{president\_of}\end{small} or \begin{small}\texttt{citizen\_of}\end{small}.

% ideally, xxxx, this motivates us to use the following hypothsis to model ....
Intuitively, entity types of the entity mention arguments pose constraints on the search space for the relation types of the relation mention (\eg, it is unlikely to find a \begin{small}\texttt{author\_of}\end{small} relation between a \begin{small}\texttt{organization}\end{small} entity and a \begin{small}\texttt{location}\end{small} entity). The proposed Hypotheses~\ref{hypo:co_occurrence} and \ref{hypo:partial_label} model types of relation mentions and entity mentions by learning an entity embedding space and a relation embedding space, respectively. The correlations between entity and relation types (and their embedding spaces) motivates us to model entity-relation interactions based on the following hypothesis.
%\vspace{-0.05cm}
\begin{hypothesis}[Entity-Relation Interaction]
\label{hypo:entity_relation}
$\newline$
For a relation mention $z=(m_1,m_2,s)$, embedding vector of $m_1$ should be a nearest neighbor of the embedding vector of $m_2$ plus the embedding vector of relation mention $z$.
\end{hypothesis}

Given the embedding vectors of any two members in \begin{small}$\{z, m_1, m_2\}$\end{small}, say \begin{small}$\bz$\end{small} and \begin{small}$\bm_1$\end{small}, Hypothesis~\ref{hypo:entity_relation} forces the ``\begin{small}$\bm_1 + \bz \approx \bm_2$\end{small}''. This helps regularize the learning of vector \begin{small}$\bm_2$\end{small} (which represents the type semantics of entity mention $m_2$) in addition to the information encoded by objective \begin{small}$O_M$\end{small} in Eq.~\eqref{eq:entity_mention_obj}.
Such a ``translating operation" between embedding vectors in a low-dimensional space has been proven effective in embedding entities and relations in a structured knowledge baes~\cite{bordes2013translating}. We extend this idea to model the type correlations (and mutual constraints) between embedding vectors of entity mentions and embedding vectors of relation mentions, which are modeled in two different low-dimensional spaces.

%%% technical idea and intuition behind the formulas
Specifically, we define error function for the triple of a relation mention and its two entity mention arguments $(z,m_1,m_2)$ using $\ell$-2 norm: \begin{small}$\tau(z)=\|\bm_1+\bz-\bm_2\|_2^2$\end{small}. A small value on \begin{small}$\tau(z)$\end{small} indicates that the embedding vectors of  $(z,m_1,m_2)$  do capture the type constraints. To enforce small errors between linkable relation mentions (in set \begin{small}$\Z_L$\end{small}) and their entity mention arguments, we use margin-based loss~\cite{bordes2013translating} to formulate a objective function as follows.
%\vspace{-0.15cm}
\begin{align}
\label{eq:entity_relation_obj}
O_{ZM}=\sum_{z_i\in\Z_L}\sum_{v=1}^V\max\big\{0, 1 + \tau(z_i) - \tau(z_v) \big\},
\end{align}
where \begin{small}$\{z_v\}_{v=1}^V$\end{small} are negative samples for \begin{small}$z$\end{small}, \ie, \begin{small}$z_v$\end{small} is randomly sampled from the negative sample set \begin{small}$\{(z',m_1,m_2)\}\cup\{(z,m'_1,m_2)\}\cup\{(z,m_1,m'_2)\}$\end{small} with \begin{small}$z'\in\Z_L$\end{small} and \begin{small}$m'\in\M_L$\end{small}~\cite{bordes2013translating}.
The intuition behind Eq.~\eqref{eq:entity_relation_obj} is simple (see also the right col. in Fig.~\ref{figure:tech_idea_illustration}): embedding vectors for a relation mention and its entity mentions are modeled in the way that, the translating error $\tau$ between them should be \textit{smaller} than the translating error of any negative sample.

\medskip\noindent
\textsf{\small\textbf{A Joint Optimization Problem.}}
% 0.2 page
Our goal is to embed all the available information for relation and entity mentions, relation and entity type labels, and text features into a $d$-dimensional entity space and a $d$-dimensional relation  space, following the three proposed hypotheses. An intuitive solution is to \textit{collectively} minimize the three objectives \begin{small}$O_Z$\end{small} \begin{small}$O_M$\end{small} and \begin{small}$O_{ZM}$\end{small}, as the embedding vectors of entity and relation mentions are shared across them. To achieve the goal, we formulate a joint optimization problem as follows.
%\vspace{-0.1cm}
\begin{align}
\label{eq:objective}
\min_{\{\bz_i\},\{\bc_j\},\{\br_k\},\{\bm_i\},\{\bc'_j\},\{\by_k\}} \O = \O_{M} + \O_{Z} + \O_{ZM}.
\end{align}
%The proposed global objective integrates corpus-level mention-feature co-occurrences, noisy type-mention associations, and interactions between relation and entity mentions. 

Optimizing the global objective $O$ in Eq.~\eqref{eq:objective} enables the learning of entity and relation embeddings to be \textit{mutually} influenced, such that, errors in each component can be constrained and corrected by the other. The joint embedding learning also helps the algorithm to find the true types for each mention, besides using text features.

In Eq.~\eqref{eq:objective}, one can also minimize the weighted combination of the three objectives \begin{small}$\{O_Z,O_M,O_{ZM}\}$\end{small} to model the importance of different signals, where weights could be  manually determined or automatically learned from data. We leave this as future work. 
%By solving the optimization problem in Eq.~\eqref{eq:objective}, we are able to represent every node in \begin{small}$G$\end{small} with a $d$-dimensional vector.
%%%%%%%%%%%%%%%%%%%%%%%
% algorithm
%%%%%%%%%%%%%%%%%%%%%%%

%#######################
\begin{algorithm}[t]
\begin{small}
\DontPrintSemicolon
\KwIn{labeled training corpus $\D_L$, text features \begin{tiny}$\{\F_z,\F_m\}$\end{tiny}, regularization parameter $\lambda$, learning rate $\alpha$, number of negative samples $V$, dim. $d$}
\KwOut{relation/entity mention embeddings \begin{tiny}$\{\bz_i\}$/$\{\bm_i\}$\end{tiny}, 
feature embeddings \begin{tiny}$\{\bc_j\},\{\bc'_j\}$\end{tiny}, relation/entity type embedding \begin{tiny}$\{\by_k\}$/$\{\br_k\}$\end{tiny}}
Initialize: vectors \begin{tiny}$\{\bz_i\}$,$\{\bm_i\}$,$\{\bc_j\}$,$\{\bc'_j\}$,$\{\by_k\}$,$\{\br_k\}$\end{tiny} as random vectors\;
\While{$\O$ in Eq.~\eqref{eq:objective} not converge}{
	\For{objective in \begin{tiny}$\{O_Z,~O_M\}$\end{tiny}}{
		 Sample a mention-feature co-occurrence $w_{ij}$; draw $V$ negative samples; update \{$\bz$, $\bc$\} based on $\L_{ZF}$, or \{$\bm$, $\bc'$\} based on $\L_{MF}$ \;
		 Sample a mention $z_i$ (or $m_i$); get its candidate types $\R_i$ (or $\Y_i$); draw $V$ negative samples; update $\bz$ and $\{\br\}$ based on \begin{tiny}$\O_Z-\L_{ZF}$\end{tiny}, or \begin{scriptsize}$\bm$\end{scriptsize} and $\{\by\}$ based on \begin{tiny}$\O_M-\L_{MF}$\end{tiny}
	} 
	Sample a relation mention \begin{tiny}$z_i=(m_1,m_2)\in\Z_L$\end{tiny}; draw $V$ negative samples; update \begin{tiny}$\{\bz_i,\bm_1,\bm_2\}$\end{tiny} based on \begin{tiny}$O_{ZM}$\end{tiny}
}
%\Return{$\{\}$ based on the estimated $\{\}$}\;
\caption{\small {Model Learning of \textsc{CoType}}}
\label{algorithm:cotype}
\end{small}
\end{algorithm}
%#######################

\subsection{Model Learning and Type Inference}
\label{subsec:algorithm}
% 0.5 page
The joint optimization problem in Eq.~\eqref{eq:objective} can be solved in multiple ways. One solution is to first learn entity mention embeddings by minimizing \begin{small}$\O_{M}$\end{small}, then apply the learned embeddings to optimize \begin{small}$\O_{MZ}+\O_{Z}$\end{small}. However, such a solution does not fully exploit the entity-relation interactions in providing mutual feedbacks between the learning of entity mention embeddings and the learning of relation mention embeddings (see \textsc{CoType-TwoStep} in Sec.~\ref{sec:experiments}). 

We design a stochastic sub-gradient descent algorithm~\cite{shalev2011pegasos} based on edge sampling strategy~\cite{tang2015line}, to efficiently solve Eq.~\eqref{eq:objective}. In each iteration, we alternatively sample from each of the three objectives \begin{small}$\{O_Z,O_M,O_{ZM}\}$\end{small} a batch of edges (\eg, \begin{small}$(z_i,f_j)$\end{small}) and their negative samples, and update each embedding vector based on the derivatives.
Algorithm~\ref{algorithm:cotype} summarizes the model learning process of \textsc{CoType}. The proof procedure in~\cite{shalev2011pegasos} can be adopted to prove convergence of the proposed algorithm (to the local minimum).
%Eq.~\eqref{eq:objective} can also be solved by a mini-batch extension of the Pegasos algorithm~\cite{shalev2011pegasos}, which is a stochastic sub-gradient descent method and thus can efficiently handle massive text corpora. Due to lack of space, we do not include derivation details here.

\smallskip
\noindent \textsf{\small \textbf{Type Inference.}}
With the learned embeddings of features and types in relation space (\ie, \begin{small}$\{\bc_i\}$, $\{\br_k\}$\end{small}) and entity space (\ie, \begin{small}$\{\bc'_i\}$, $\{\by_k\}$\end{small}), we can perform nearest neighbor search in the target relation type set \begin{small}$\R$\end{small}, or a top-down search on the target entity type hierarchy \begin{small}$\Y$\end{small}, to estimate the relation type (or the entity type-path) for each (unlinkable) test relation mention \begin{small}$z\in\Z_U$\end{small} (test entity mention \begin{small}$m\in\M\setminus\M_L$\end{small}). Specifically, on the entity type hierarchy, we start from the tree's root and recursively find the best type among the children types by measuring the \textit{cosine similarity} between entity type embedding and the vector representation of $m$ in our learned entity embedding space. By extracting text features from $m$'s local context (denoted by set \begin{small}$\F_m(m)$\end{small}), we represent $m$ in the learned entity embedding space using the vector \begin{small}$\bm = \sum_{f_j\in\F_m(m)}\bc'_{j}$\end{small}.  Similarly, for test relation mention $z$, we represent it in our learned relation embedding space by \begin{small}$\bz = \sum_{f_j\in\F_z(z)}\bc_{j}$\end{small} where \begin{small}$\F_z(z)$\end{small} is the set of text features extracted from $z$'s local context $s$. The search process stops when we reach to a leaf type on the type hierarchy, or the similarity score is below a pre-defined threshold $\eta>0$. If the search process returns an empty type-path (or type set), we output the predicted type label as \begin{small}\texttt{None}\end{small} for the mention.

\smallskip
\noindent
\textsf{\small \textbf{Computational Complexity Analysis.}}
%%% graph construction
%Feature generation in \textsc{CoType} takes \begin{small}$O(|\D|$)\end{small} time where \begin{small}$|\D|$\end{small} denotes the number of tokens in corpus \begin{small}$\D$\end{small}.
%%% PLE
Let \begin{small}$E$\end{small} be the total number of objects in \textsc{CoType} (entity and relation mentions, text features and type labels).
By alias table method~\cite{tang2015line}, setting up alias tables takes \begin{small}$O(E)$\end{small} time for all the objects, and sampling a negative example takes constant time. 
%%%
In each iteration of Algorithm~\ref{algorithm:cotype}, optimization with negative sampling (\ie, optimizing second-order proximity and translating objective) takes \begin{small}$O(dV)$\end{small}, and optimization with partial-label loss takes \begin{small}$O\big(dV(|\R|+|\Y|)\big)$\end{small} time. Similar to~\cite{tang2015line}, we find the number of iterations for Algorithm~\ref{algorithm:cotype} to converge is usually \textit{proportional to} the number of object interactions extracted from \begin{small}$\D$\end{small} (\eg, unique mention-feature pairs and mention-type associations), denoted as \begin{small}$R$\end{small}. Therefore, the overall time complexity of \textsc{CoType} is \begin{small}$O\big(dRV(|\R|+|\Y|)\big)$\end{small} (as \begin{small}$R\geq E$\end{small}), which is \textit{linear} to the total number of object interactions \begin{small}$R$\end{small} in the corpus.

%%%%%%%%%%%%%%%%%%%%%%%
% experiments
%%%%%%%%%%%%%%%%%%%%%%%
%\vspace{-0.15cm}
\section{Experiments}
\label{sec:experiments}
%Using three text corpora of different domains, we test the proposed framework on entity recognition and typing, relation classification, relation extraction, and scalability.

\begin{table}[t]
\vspace{-0.3cm}
\begin{scriptsize}
\begin{center}
\begin{tabularx}{0.87\linewidth}{l lll}
\hline
\textbf{Data sets~~~} & \textbf{NYT} & \textbf{Wiki-KBP} & \textbf{BioInfer} \\
\hline
$\#$Relation/entity types & 24~/~47  & 19~/~126 & 94~/~2,200 \\ %\hline
$\#$Documents (in $\D$) & 294,977~ & 780,549 & 101,530 \\ %\hline
$\#$Sentences (in $\D$) & 1.18M & 1.51M & 521k \\ %\hline
$\#$Training RMs (in $\Z_L$) & 353k & 148k & 28k \\ %\hline
$\#$Training EMs (in $\Z_L$) & 701k & 247k  & 53k \\ %\hline
$\#$Text features (from $\D_L$) & 2.6M & 1.3M & 575k \\ %\hline
$\#$Test Sentences (from $\Z_U$) & 395 & 448 & 708 \\ %\hline
$\#$Ground-truth RMs & 3,880 & 2,948 & 3,859 \\ %\hline
$\#$Ground-truth EMs & 1,361 & 1,285 & 2,389 \\ %\hline
\hline
\end{tabularx}
%\vspace{-0.2cm}
\caption{Statistics of the datasets in our experiments.}
\label{table:data_stats}
\vspace{-0.4cm}
\end{center}
\end{scriptsize}
\end{table}

%\vspace{-0.1cm}
\subsection{Data Preparation and Experiment Setting}
% \subsection{Data Preparation}
\label{subsec:data_preparation}
Our experiments use three public datasets\footnote{\small  Codes and datasets used in this paper can be downloaded at: \url{https://github.com/shanzhenren/CoType}.} from different domains. (1) \textbf{NYT}~\cite{riedel2010modeling}: The training corpus consists of 1.18M sentences sampled from $\sim$294k 1987-2007 New York Times news articles. 395 sentences are manually annotated by authors of \cite{hoffmann2011multiR} to form the test data;
(2) \textbf{Wiki-KBP}~\cite{ling2012fine}: It uses 1.5M sentences sampled from $\sim$780k Wikipedia articles~\cite{ling2012fine} as training corpus and 14k manually annotated sentences from 2013 KBP slot filling assessment results~\cite{ellislinguistic} as test data.
(3) \textbf{BioInfer}~\cite{pyysalo2007bioinfer}: It consists of 1,530 manually annotated biomedical paper abstracts as test data and 100k sampled PubMed paper abstracts as training corpus.
Statistics of the datasets are shown in Table~\ref{table:data_stats}.

\smallskip
\noindent
\textsf{\small\textbf{Automatically Labeled Training Corpora.}}
%% distant supervisions: entity mentions, candidate labels, target types
The NYT training corpus has been heuristically labeled using distant supervision following the procedure in~\cite{riedel2010modeling}.
For Wiki-KBP and BioInfer training corpora, we utilized DBpedia Spotlight\footnote{\scriptsize \url{http://spotlight.dbpedia.org/}}, a state-of-the-art entity disambiguation tool, to map the detected entity mentions \begin{small}$\M$\end{small} to Freebase entities.
We then followed the procedure introduced in Secs.~\ref{sec:problem} and \ref{subsec:candidate_generation} to obtain candidate entity and relation types, and constructed the training data \begin{small}$\D_L$\end{small}.
For target types, we discard the relation/entity types which cannot be mapped to Freebase from the test data while keeping the Freebase entity/relation types (not found in test data) in the training data (see Table~\ref{table:data_stats} for the type statistics).

%We adopted the type mapping created by Grillick \etal~\cite{gillick2014context} to obtain target types for each mention in OntoNotes dataset.
%For BBN dataset, we manually created a type mapping for its target types. Among the original 93 types in BBN, 22 non-entity types (\eg, \texttt{date}) and 12 descriptor types (\eg, \texttt{xxx}) cannot be mapped to Freebase types. We discarded these types from the dataset and removed test mentions of which types were all discarded.
%In particular, we discarded types which cannot be mapped to Freebase types in BBN dataset (47 out of 93).

\smallskip
\noindent
\textsf{\small\textbf{Feature Generation.}}
Table~\ref{table:features} lists the set of text features of relation mentions used in our experiments. We followed~\cite{ling2012fine} to generate text features for entity mentions. Dependency parse-based features were excluded as only POS-tagged corpus is given as input. 
We used a 6-word window to extract context features for each mention (3 words on the left and the right).
We applied the Stanford CoreNLP tool~\cite{manning2014stanford} to get POS tags.
Brown clusters were derived for each corpus using public implementation\footnote{\scriptsize \url{https://github.com/percyliang/brown-cluster}}.
%% details on pruning the features;
The same kinds of features were used in all the compared methods in our experiments.

\smallskip
\noindent
\textsf{\small\textbf{Evaluation Sets.}}
%% how to form the evalaution sets for two different tasks.
For all three datasets, we used the provided training/test set partitions of the corpora. In each dataset, relation mentions in sentences are manually annotated with their relation types and the entity mention arguments are labeled with entity type-paths (see Table~\ref{table:data_stats} for the statistics of test data). 
%% validation set
We further created a \textit{validation set} by randomly sampling 10\% mentions from each test set and used the remaining part to form the \textit{evaluation set}.

\smallskip\noindent
\textsf{\small\textbf{Compared Methods.}}
We compared \textsc{CoType} with its variants which model parts of the proposed hypotheses. Several state-of-the-art relation extraction methods (\eg, supervised, embedding, neural network) were also implemented (or tested using their published codes):
(1) \textbf{\small DS+Perceptron}~\cite{ling2012fine}: adopts multi-label learning on automatically labeled training data \begin{small}$\D_L$\end{small}.
(2) \textbf{\small DS+Kernel}~\cite{mooney2005subsequence}: applies bag-of-feature kernel~\cite{mooney2005subsequence} to train a SVM classifier using \begin{small}$\D_L$\end{small};
(3) \textbf{\small DS+Logistic}~\cite{mintz2009distant}: trains a multi-class logistic classifier\footnote{\scriptsize  We use liblinear package from \url{https://github.com/cjlin1/liblinear}} on \begin{small}$\D_L$\end{small};
(4) \textbf{\small DeepWalk}~\cite{perozzi2014deepwalk}: embeds mention-feature co-occurrences and mention-type associations as a homogeneous network (with binary edges);
(5) \textbf{\small LINE}~\cite{tang2015line}:
uses second-order proximity model with edge sampling on a feature-type bipartite graph (where edge weight $w_{jk}$ is the number of relation mentions having feature $f_j$ and type $r_k$);
(6) \textbf{\small MultiR}~\cite{hoffmann2011multiR}: is a state-of-the-art distant supervision method, which models noisy label in \begin{small}$\D_L$\end{small} by multi-instance multi-label learning;
(7) \textbf{\small FCM}~\cite{gormley2015improved}: adopts neural language model to perform compositional embedding;
(8) \textbf{\small DS-Joint}~\cite{li2014incremental}: jointly extract entity and relation mentions using structured perceptron on human-annotated sentences. We used \begin{small}$\D_L$\end{small} to train the model.

For \textsc{CoType}, besides the proposed model, \textbf{\small CoType}, we compare (1) \textbf{\small CoType-RM}: This variant only optimize objective \begin{small}$O_{Z}$\end{small} to learning feature and type embeddings for relation mentions; and (2) \textbf{\small CoType-TwoStep}: It first optimizes \begin{small}$\O_M$\end{small}, then use the learned entity mention embedding \begin{small}$\{\bm_i\}$\end{small} to initialize the minimization of \begin{small}$O_{Z}+O_{ZM}$\end{small}---it represents a ``pipeline" extraction diagram.

%%% entity recognition and typing
To test the performance on entity recognition and typing, we also compare with several entity recognition systems, including a supervised method \textbf{\small HYENA}~\cite{yosef2012hyena}, distant supervision methods (\textbf{\small FIGER}~\cite{ling2012fine}, \textbf{\small Google}~\cite{gillick2014context}, \textbf{\small WSABIE}~\cite{Yogatama2015embedding}), and a noise-robust approach \textbf{\small PLE}~\cite{ren2016label}.

\nop{
(1) \textbf{\small FIGER}~\cite{ling2012fine}: perform fine-grained entity recognition using multi-label perceptron; 
(2) \textbf{\small Google}~\cite{Yogatama2015embedding}: applies logistic classifier for entity typing; 
(3) textbf{\small HYENA}~\cite{Yogatama2015embedding}: trains a hierarchical SVM classifier; 
(4) \textbf{\small WSABIE}~\cite{Yogatama2015embedding}: adopts WARP loss with kernel extension to learn embeddings of features and types;
(5) \textbf{\small PLE}~\cite{ren2016label}: models noisy type labels and type correlation by heterogeneous graph-based embedding.
}

\begin{table}
\begin{scriptsize}
\vspace{-0.4cm}
\begin{center}
\hspace*{-0.3cm}
\begin{tabularx}{1.05\linewidth}{b | aaa | aaa | aaa }
\hline
 & \multicolumn{3}{c|}{\textbf{NYT}} &  \multicolumn{3}{c|}{\textbf{Wiki-KBP}} & \multicolumn{3}{c}{\textbf{BioInfer}}\\
\textbf{Method}
& \textsf{S-F1} & \textsf{Ma-F1}  & \textsf{Mi-F1} 
& \textsf{S-F1} & \textsf{Ma-F1} & \textsf{Mi-F1} 
& \textsf{S-F1} & \textsf{Ma-F1}  & \textsf{Mi-F1} 
\\ \hline
FIGER~\cite{ling2012fine}
% NYT
& 0.40 & 0.51 & 0.46
% Wiki-KBP
& 0.29 & 0.56 & 0.54
% Biomedical
& 0.69 & 0.71 & 0.71
\\
Google~\cite{gillick2014context}
% NYT
& 0.38 & 0.57 & 0.52
% Wiki-KBP
& 0.30 & 0.50 & 0.38
% Biomedical
& 0.69 & 0.72 & 0.65
\\
HYENA~\cite{yosef2012hyena}
% NYT
& 0.44 & 0.49 & 0.50
% Wiki-KBP
& 0.26 & 0.43 & 0.39
% Biomedical
& 0.52 & 0.54 & 0.56
\\
DeepWalk\cite{perozzi2014deepwalk}
% NYT
& 0.49 & 0.54 & 0.53
% Wiki-KBP
& 0.21 & 0.42 & 0.39
% Biomedical
& 0.58 & 0.59 & 0.61
\\
WSABIE\cite{Yogatama2015embedding}
% NYT
& 0.53 & 0.57 & 0.58
% Wiki-KBP
& 0.35 & 0.55 & 0.50
% Biomedical
& 0.64 & 0.66 & 0.65
\\
PLE~\cite{ren2016label}
% NYT
& 0.56 & 0.60 & 0.61
% Wiki-KBP
& 0.37 & 0.57 & 0.53
% Biomedical
& 0.70 & 0.71 & 0.72
\\
CoType
% NYT
& \textbf{0.60} & \textbf{0.65} & \textbf{0.66}
% Wiki-KBP
& \textbf{0.39} & \textbf{0.61} & \textbf{0.57}
% Biomedical
& \textbf{0.74} & \textbf{0.76} & \textbf{0.75}
\\\hline
\end{tabularx}
%\vspace{-0.25cm}
\caption{Performance comparison of entity recognition and typing  (using strict, micro and macro metrics~\cite{ling2012fine}) on the three datasets.}
\label{table:entity_recognition}
\end{center}
\vspace{-0.4cm}
\end{scriptsize}
\end{table}

\smallskip
\noindent
\textsf{\small \textbf{Parameter Settings.}}
%% PLE setting: alpha, lambda, lr, # iters.
In our testing of \textsc{CoType} and its variants, we set \begin{small}$\alpha=0.025$\end{small}, \begin{small}$\eta=0.35$\end{small} and \begin{small}$\lambda=10^{-4}$\end{small} based on the analysis on validation sets. For convergence criterion, we stopped the loop in Algorithm~\ref{algorithm:cotype} if the relative change of \begin{small}$\O$\end{small} in Eq.~\eqref{eq:objective} is smaller than \begin{small}$10^{-4}$\end{small}.
%% embedding dimensionality, # negative samples
For fair comparison, the dimensionality of embeddings $d$ was set to $50$ and the number of negative samples \begin{small}$V$\end{small} was set to $5$ for all embedding methods, as used in \cite{tang2015line}. For other tuning parameters in the compared methods, we tuned them on validation sets and picked the values which lead to the best performance.
%% DeepWalk setting
%For DeepWalk, we set window size as 10, walk length as 40, walks per vertex as 40, as used in~\cite{perozzi2014deepwalk}.
%% learning rate of LINE, PTE,
%Learning rates of LINE and PTE were set to \begin{small}$\rho_t=\rho_0(1-t/T)$\end{small} with \begin{small}$\rho_0=0.025$\end{small} where $T$ is total number of edge samples (set to 10 times of the number of edges), as used in \cite{tang2015pte} and \cite{tang2015line}.
%After tuning on validation sets, we set learning rate as $ 0.001$ for WSABIE, and set the regularization parameters in PL-SVM and CLPL as $0.1$.

\begin{table}[t]
\vspace{-0.3cm}
\begin{scriptsize}
\begin{center}
%\vspace{0.0cm}
\begin{tabularx}{0.8\linewidth}{l | ccc}
\hline
\textbf{Method} & \textbf{~NYT~~} & \textbf{Wiki-KBP} & \textbf{BioInfer} \\
\hline
DS+Perceptron~\cite{ling2012fine}~~~~ & 0.641 & 0.543 & 0.470 \\ %\hline
DS+Kernel~\cite{mooney2005subsequence} & 0.632 & 0.535 & 0.419 \\ %\hline
DeepWalk~\cite{perozzi2014deepwalk} & 0.580 & 0.613 & 0.408 \\ %\hline
LINE~\cite{tang2015line} & 0.765 & 0.617 & 0.557 \\ %\hline
DS+Logistic~\cite{mintz2009distant} & 0.771 & 0.646 & 0.543 \\ %\hline
MultiR~\cite{hoffmann2011multiR} & 0.693 & 0.633 & 0.501 \\
FCM~\cite{gormley2015improved} & 0.688 & 0.617 & 0.467 \\ \hline
CoType-RM & 0.812 & 0.634 & 0.587 \\
CoType-TwoStep & 0.829 & 0.645 & 0.591 \\
CoType & \textbf{0.851} & \textbf{0.669} & \textbf{0.617} \\
\hline
\end{tabularx}
%\vspace{-0.2cm}
\caption{Performance comparison on relation classification accuracy over ground-truth relation mentions on the three datasets.}
\label{table:relation_classification}
\vspace{-0.4cm}
\end{center}
\end{scriptsize}
\end{table}

\smallskip\noindent
\textsf{\small \textbf{Evaluation Metrics.}}
For entity recognition and typing, we to use strict, micro, and macro F1 scores, as used in~\cite{ling2012fine}, for evaluating both detected entity mention boundaries and predicted entity types.
We consider two settings in evaluation of relation extraction. For relation classification, ground-truth relation mentions are given and \begin{small}\texttt{None}\end{small} label is excluded. We focus on testing type classification accuracy.
For relation extraction, we adopt standard Precision (\begin{small}\textbf{P}\end{small}), Recall (\begin{small}\textbf{R}\end{small}) and F1 score~\cite{mooney2005subsequence,bach2007review}. Note that all our evaluations are \textit{sentence-level} (\ie, context-dependent), as discussed in~\cite{hoffmann2011multiR}.

%\vspace{-0.0cm}
% \subsection{Performance Comparison}
\subsection{Experiments and Performance Study}
\label{subsec:performance_comparison}
\noindent
\textsf{\small\textbf{1. Performance on Entity Recognition and Typing.}}
Among the compared methods, only FIGER~\cite{ling2012fine} can detect entity mention. We apply our detection results (\ie, \begin{small}$\M$\end{small}) as input for other methods. Table~\ref{table:entity_recognition} summarizes the comparison results on the three datasets. Overall, \textsc{CoType} outperforms others on all metrics on all three datasets (\eg, it obtains a 8\% improvement on Micro-F1 over the \textit{next best method} on NYT dataset). Such performance gains mainly come from (1) a more robust way of modeling noisy candidate types (as compared to supervised method and distant supervision methods which ignore label noise issue); and (2) the joint embedding of entity and relation mentions in a mutually enhancing way (vs. the noise-robust method PLE~\cite{ren2016label}). This demonstrates the effectiveness of enforcing Hypothesis~\ref{hypo:entity_relation} in \textsc{CoType} framework.

\begin{table*}
\begin{scriptsize}
\vspace{-0.3cm}
%\vspace{0.0cm}
\begin{center}
\begin{tabularx}{\textwidth}{  l | aaaa | aaaa | aaaa }
\hline
 & \multicolumn{4}{c|}{\textbf{NYT}~\cite{riedel2010modeling,hoffmann2011multiR}} &  \multicolumn{4}{c|}{\textbf{Wiki-KBP}~\cite{ellislinguistic,ling2012fine}} & \multicolumn{4}{c}{\textbf{BioInfer}~\cite{pyysalo2007bioinfer}}\\
\textbf{Method}
& \textbf{Prec} & \textbf{Rec}  & \textbf{F1} & \textbf{Time}
& \textbf{Prec} & \textbf{Rec} & \textbf{F1} & \textbf{Time}
& \textbf{Prec} & \textbf{Rec}  & \textbf{F1} & \textbf{Time}
\\ \hline
DS+Perceptron~\cite{ling2012fine}
% NYT
& 0.068 & \textbf{0.641} & 0.123 & 15min
% Wiki-KBP
& 0.233 & 0.457 & 0.308 & 7.7min
% Biomedical
& 0.357 & 0.279 & 0.313 & 3.3min
\\
DS+Kernel~\cite{mooney2005subsequence} 
% NYT
& 0.095 & 0.490 & 0.158 & 56hr
% Wiki-KBP
& 0.108 & 0.239 & 0.149 & 9.8hr
% Biomedical
& 0.333 & 0.011 & 0.021 & 4.2hr
\\
DS+Logistic~\cite{mintz2009distant}
% NYT
& 0.258 & 0.393 & 0.311 & 25min
% Wiki-KBP
& 0.296 & 0.387 & 0.335 & 14min
% Biomedical
& \textbf{0.572} & 0.255 & 0.353 & 7.4min
\\
DeepWalk~\cite{perozzi2014deepwalk}
% NYT
& 0.176 & 0.224 & 0.197 & 1.1hr
% Wiki-KBP
& 0.101 & 0.296 & 0.150 & 27min
% Biomedical
& 0.370 & 0.058 & 0.101 & 8.4min
\\
LINE~\cite{tang2015line}
% NYT
& 0.335 & 0.329 & 0.332 & 2.3min
% Wiki-KBP
& 0.360 & 0.257 & 0.299 & 1.5min
% Biomedical
& 0.360 & 0.275 & 0.312 & 35sec
\\ 
MultiR~\cite{hoffmann2011multiR}
% NYT
& 0.338 & 0.327 & 0.333 & 5.8min
% Wiki-KBP
& 0.325 & 0.278 & 0.301 & 4.1min
% Biomedical
& 0.459 & 0.221 & 0.298 & 2.4min
\\
FCM~\cite{gormley2015improved}
% NYT
& 0.553 & 0.154 & 0.240 & 1.3hr
% Wiki-KBP
& 0.151 & \textbf{0.500} & 0.301 & 25min
% Biomedical
& 0.535 & 0.168 & 0.255 & 9.7min
\\
DS-Joint~\cite{li2014incremental}
% NYT
& \textbf{0.574} & 0.256 & 0.354 & 22hr
% Wiki-KBP
& \textbf{0.444} & 0.043 & 0.078 & 54hr
% Biomedical
& 0.102 & 0.001 & 0.002 & 3.4hr
\\
\hline
CoType-RM
% NYT
& 0.467 & 0.380 & 0.419 & 2.6min
% Wiki-KBP
& 0.342 & 0.339 & 0.340 & 1.5min
% Biomedical
& 0.482 & 0.406 & 0.440 & 42sec
\\
CoType-TwoStep
% NYT
& 0.368 & 0.446 & 0.404 & 9.6min
% Wiki-KBP
& 0.347 & 0.351 & 0.349 & 6.1min
% Biomedical
& 0.502 & 0.405 & 0.448 & 3.1min
\\
CoType
% NYT
& 0.423 & 0.511 & \textbf{0.463} & 4.1min
% Wiki-KBP
& 0.348 & 0.406 & \textbf{0.369} & 2.5min
% Biomedical
& 0.536 & \textbf{0.424} & \textbf{0.474} & 78sec
\\\hline
\end{tabularx}
\vspace{-0.1cm}
\caption{Performance comparison on end-to-end relation extraction (at the highest F1 point) on the three datasets.}
\label{table:relation_extraction}
\end{center}
\vspace{-0.2cm}
\end{scriptsize}
\end{table*}

\smallskip
\noindent
\textsf{\small\textbf{2. Performance on Relation Classification.}}
To test the effectiveness of the learned embeddings in representing type semantics of relation mentions,  we compare with other methods on classifying the ground-truth relation mention in the evaluation set by target types \begin{small}$\R$\end{small}. Table~\ref{table:relation_classification} summarizes the classification accuracy. \textsc{CoType} achieves superior accuracy compared to all other methods and variants (\eg, obtains over 10\% enhancement on both the NYT and BioInfer datasets over the next best method).
All compared methods (except for MultiR) simply treat \begin{small}$\D_L$\end{small} as ``perfectly labeled'' when training models. The improvement of \textsc{CoType-RM} validates the importance on careful modeling of label noise (\ie, Hypothesis~\ref{hypo:partial_label}). Comparing \textsc{CoType-RM} with MultiR, superior performance of \textsc{CoType-RM} demonstrates the effectiveness of partial-label loss over multi-instance learning. Finally, \textsc{CoType} outperforms \textsc{CoType-RM} and \textsc{CoType-TwoStep} validates that the propose translation-based embedding objective is effective in capturing entity-relation cross-constraints.

\begin{figure}
\centering
\vspace{-0.4cm}
\subfigure[{NYT}]{
\includegraphics[width = 40 mm]{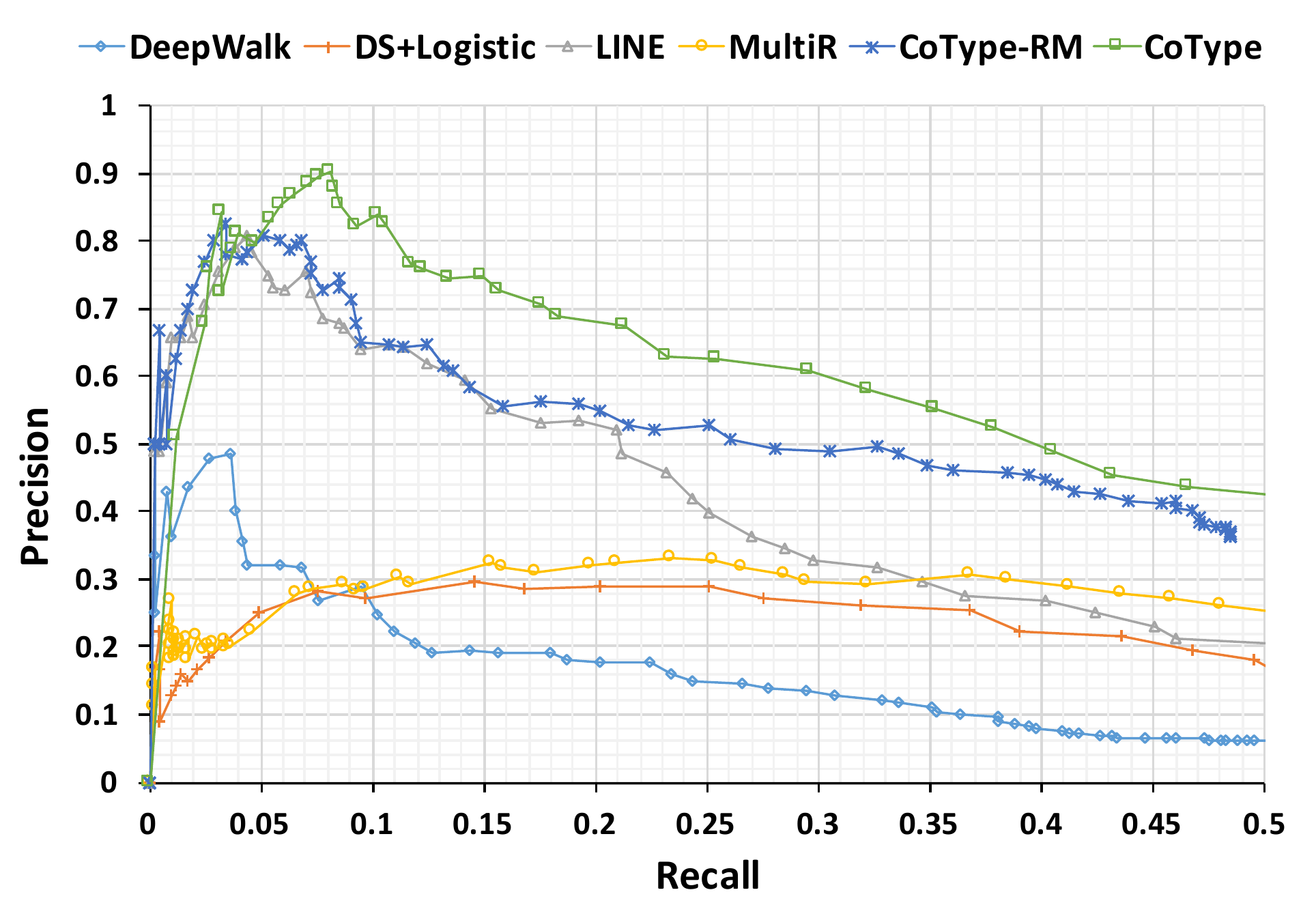}
}
%\subfigure[{Wiki-KBP}]{
%\includegraphics[width = 50 mm]{figures/pr_nyt.pdf}
%}
\subfigure[{BioInfer}]{
\includegraphics[width = 40 mm]{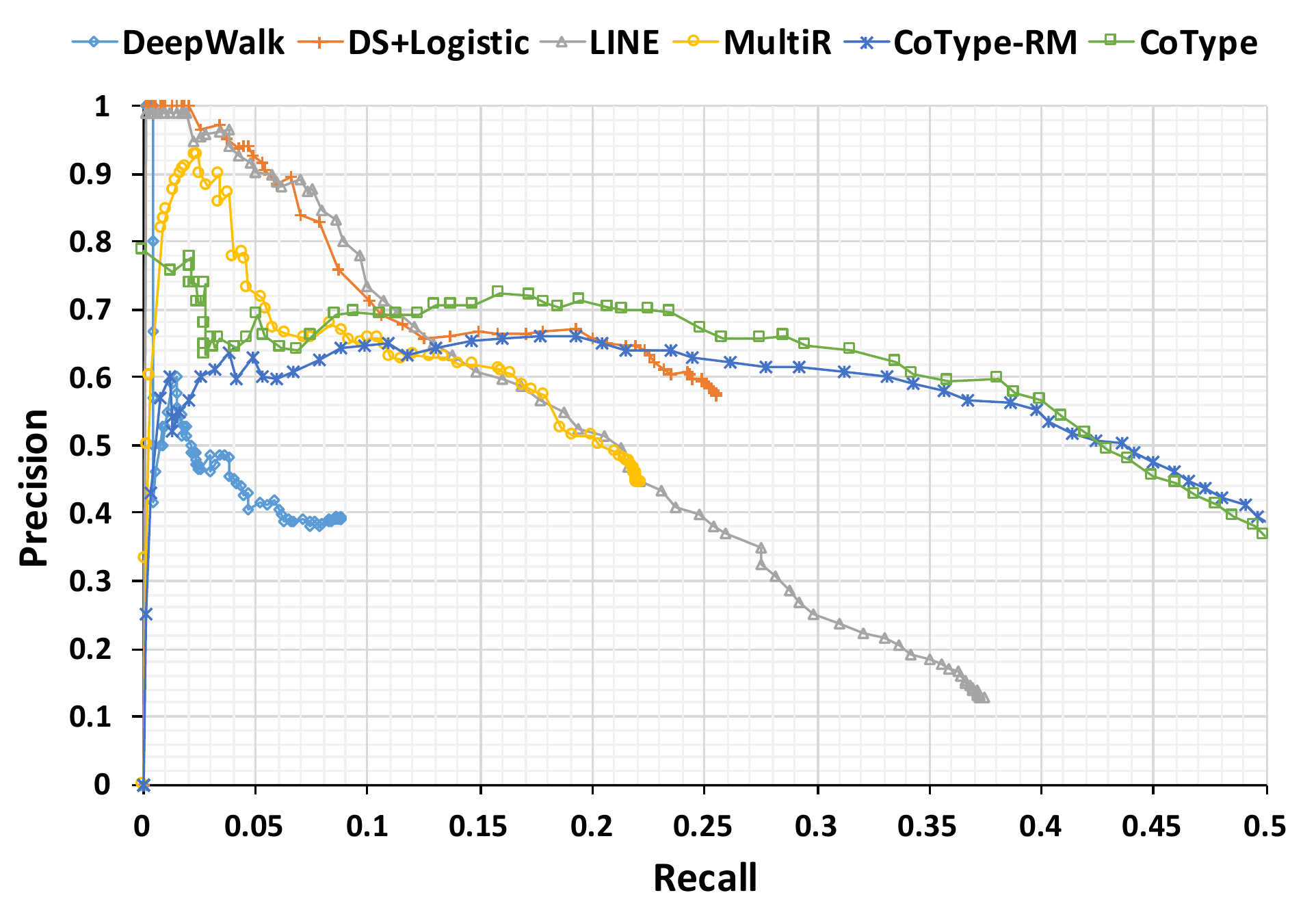}
}
\vspace{-0.3cm}
\caption{Precision-recall curves of relation extraction on NYT and BioInfer datasets. Similar trend is also observed on the Wiki-KBP dataset.}
\label{figure:precision_recall_curve}
\vspace{-0.1cm}
\end{figure}

\smallskip\noindent
\textsf{\small\textbf{3. Performance on Relation Extraction.}}
To test the domain independence of \textsc{\small CoType} framework, we conduct evaluations on the end-to-end relation extraction. As only MultiR and DS-Joint are able to detection entity and relation mentions in their own framework, we apply our detection results to other compared methods.
Table~\ref{table:relation_extraction} shows the evaluation results  as well as runtime of different methods. In particular, results at each method's highest F1 score point are reported, after tuning the threshold for each method for determining whether a test mention is \begin{small}\texttt{None}\end{small} or some target type.
Overall, \textsc{\small CoType} outperforms all other methods on F1 score on all three datasets. We observe that DS-Joint and MultiR suffer from low recall, since their entity detection modules do not work well on \begin{small}$\D_L$\end{small} (where many tokens have false negative tags). This demonstrates the effectiveness of the proposed domain-agnostic text segmentation algorithm (see Sec.~\ref{subsec:candidate_generation}).
%Superior performance of \textsc{CoType-RM} over the distant supervision methods shows that modeling noisy candidate types using partial-label loss benefits both precision and recall, as it enables more accurate learning of feature embeddings. 
We found that the incremental diagram of learning embedding (\ie, \begin{small}\textsc{CoType-TwoStep}\end{small}) brings only marginal improvement. In contrast, \textsc{\small CoType} adopts a ``joint modeling" diagram following Hypothesis~\ref{hypo:entity_relation} and achieves significant improvement. In Fig.~\ref{figure:precision_recall_curve}, precision-recall curves on NYT and BioInfer datasets further show that \textsc{\small CoType} can still achieve descent precision with good recall preserved. 

\begin{figure}[h]
\centering
%\vspace{-0.2cm}
\subfigure[Scalability of \textsc{CoType}]{
\includegraphics[width = 40 mm]{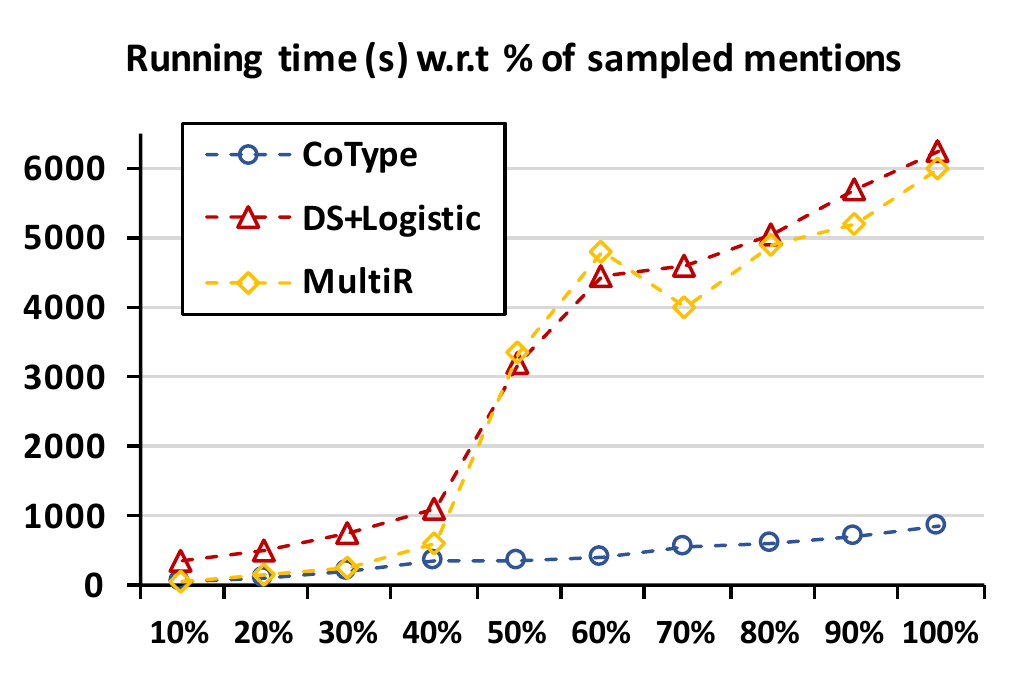}
\label{figure:scalability}
}
\subfigure[Effect of training set size]{
\includegraphics[width = 40 mm]{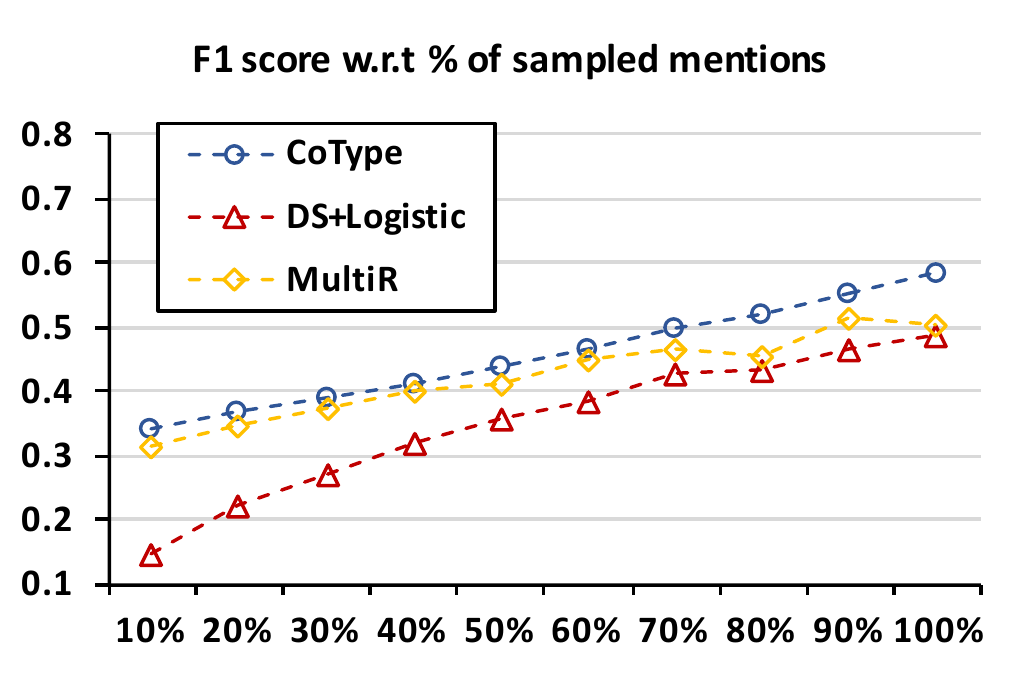}
\label{figure:train_size_effect}
}
\vspace{-0.2cm}
\caption{(a) Scalability study on \textsc{CoType} and the compared methods; and (b) Performance changes of relation extraction with respect to sampling ratio of relation mentions on the \textbf{Bioinfer} dataset.}
%\vspace{-0.5cm}
\end{figure}

\smallskip
\noindent
\textsf{\small\textbf{4. Scalability.}}
In addition to the runtime shown in Table~\ref{table:relation_extraction}, Fig.~\ref{figure:scalability} tests the scalability of \textsc{CoType} compared with other methods, by running on BioInfer corpora sampled using different ratios.
CoType demonstrates a linear runtime trend (which validates our time complexity in Sec.~\ref{subsec:algorithm}), and is the only method that is capable of processing the full-size dataset without significant time cost.

%\vspace{-0.0cm}
\subsection{Case Study}
\noindent
\textsf{\small\textbf{1.~Example output on news articles.}}
Table~\ref{table:example_output} shows the output of \textsc{\small CoType}, MultiR and Logistic on two news sentences from the Wiki-KBP dataset. CoType extracts more relation mentions (\eg, \texttt{\small children}), and predict entity/relation types with better accuracy. Also, \textsc{\small CoType} can jointly extract typed entity and relation mentions while other methods cannot (or need to do it incrementally).

\smallskip\noindent
\textsf{\small\textbf{2.~Testing the effect of training corpus size.}}
Fig.~\ref{figure:train_size_effect} shows the performance trend on Bioinfer dataset when varying the sampling ratio (subset of relation mentions randomly sampled from the training set). F1 scores of all three methods improves as the sampling ratio increases. CoType performs best in all cases, which demonstrates its robust performance across corpora of various size.

\begin{figure}[t]
\vspace{-0.15cm}
\includegraphics[width = 82 mm]{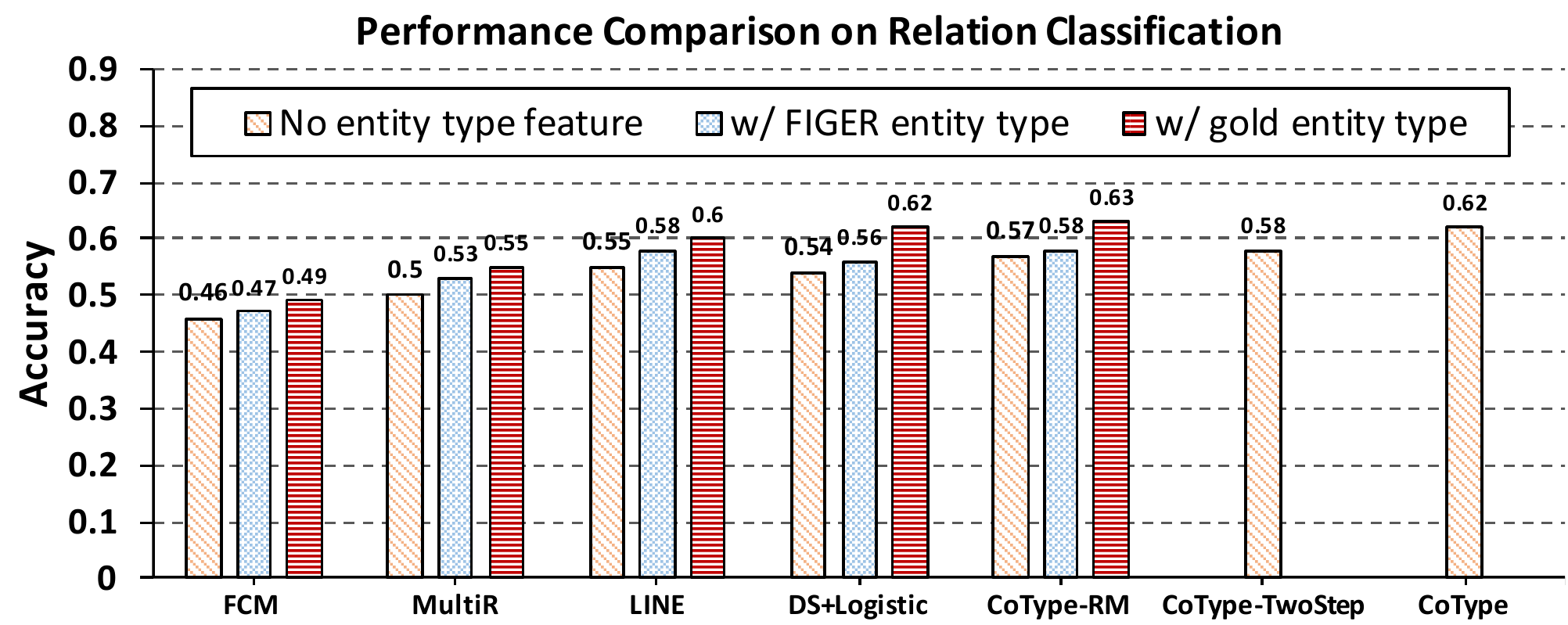}
\vspace{-0.1cm}
\caption{Study of entity type error propagation on the \textbf{BioInfer} dataset.}
\label{figure:pipeline_system}
%\vspace{-0.4cm}
\end{figure}

\smallskip\noindent
\textsf{\small\textbf{3.~Study the effect of entity type error in relation classification.}}
To investigate the ``error propagation'' issue of incremental pipeline, we test the changes of relation classification performance by (1) training models without entity types as features; (2) using entity types predicted by FIGER~\cite{ling2012fine} as features; and (3) using ground-truth (``perfect'') entity types as features. 
Fig.~\ref{figure:pipeline_system} summarize the accuracy of \textsc{\small CoType}, its variants and the compared methods.
We observe only marginal improvement when using FIGER-predicted types but significant improvement when using ground-truth entity types---this validates the error propagation issue. Moreover, we find that \textsc{\small CoType} achieves an accuracy close to that of the next best method (\ie, \begin{small}
DS + Logistic + Gold entity type\end{small}). This demonstrates the effectiveness of our proposed joint entity and relation embedding.

%%%%%%%%%%%%%%%%%%%%%%%
% related
%%%%%%%%%%%%%%%%%%%%%%%
%\vspace{-0.15cm}
\section{Related Work}
\label{sec:related}
%%% 0.5 page

\noindent
\textbf{\small\textsf{Entity and Relation Extraction.}}
%% incremental pipeline extraction methods
There have been extensive studies on extracting typed entities and relations in text (\ie, context-dependent extraction). Most existing work follows an \textit{incremental} diagram---they first perform entity recognition and typing~\cite{nadeau2007survey,ratinov2009design} to extract typed entity mentions, and then solve relation extraction~\cite{bach2007review,guodong2005exploring} to identify relation mentions of target types. Work along both lines can be categorized in terms of the degree of supervision. While supervised entity recognition systems~\cite{finkel2005incorporating,nadeau2007survey} focus on a few common entity types, weakly-supervised methods~\cite{gupta14evalpatterns,nakashole2013fine} and distantly-supervised methods~\cite{ren2015clustype,Yogatama2015embedding,ling2012fine} use large text corpus and a small set of seeds (or a knowledge base) to induce patterns or to train models, and thus can apply to different domains without additional human annotation labor.
For relation extraction, similarly, weak supervision~\cite{bunescu2007learning,etzioni2004web} and distant supervision~\cite{nagesh2014noisy,xu2013filling,surdeanu2012MIME,hoffmann2011multiR,riedel2010modeling,mintz2009distant} approaches are proposed to address the domain restriction issue in traditional supervised systems~\cite{bach2007review,mooney2005subsequence,guodong2005exploring}. 
However, such a ``pipeline" diagram ignores the dependencies between different sub tasks and may suffer from error propagation between the tasks. 

%% joint extraction methods
Recent studies try to integrate entity extraction with relation extraction by performing global sequence labeling for both entities and relations~\cite{li2014incremental,miwa2014modeling,augenstein2015extracting}, incorporating type constraints between relations and their arguments~\cite{roth2007global}, or modeling factor graphs~\cite{singh2013joint}. However, these methods require human-annotated corpora (cleaned and general) for model training and rely on existing entity detectors to provide entity mentions.
 %% our differences and advantages
By contrast, the \textsc{CoType} framework runs domain-agnostic segmentation algorithm to mine entity mentions and adopts a label noise-robust objective to train models using distant supervision. 
In particular, \cite{augenstein2015extracting} integrates entity classification with relation extraction using distant supervision but it ignores label noise issue in the automatically labeled training corpora.

\textsc{CoType} combines the best of two worlds---it leverages the noisy distant supervision in a robust way to address domain restriction (vs. existing joint extraction methods~\cite{li2014incremental,miwa2014modeling}), and models entity-relation interactions jointly with other signals to resolve error propagation (vs. current distant supervision methods~\cite{surdeanu2012MIME,mintz2009distant}).

\smallskip\noindent
\textbf{\small\textsf{Learning Embeddings and Noisy Labels.}}
%% text and network embedding
Our proposed framework incorporates embedding techniques used in modeling words and phrases in large text corpora~\cite{mikolov2013distributed,Yogatama2015embedding,salehi2015word} ,and nodes and links in graphs/networks~\cite{tang2015line,perozzi2014deepwalk}. Theses methods assume links are all correct (in unsupervised setting) or labels are all true (in supervised setting). \textsc{CoType} seeks to \textit{model the true links and labels} in the embedding process (\eg, see our comparisons with LINE~\cite{tang2015line}, DeepWalk~\cite{perozzi2014deepwalk} and FCM~\cite{gormley2015improved} in Sec.~\ref{subsec:performance_comparison}).
Different from embedding structured KB entities and relations~\cite{bordes2013translating,toutanova2015representing}, our task focuses on embedding entity and relation mentions in \textit{unstructured} contexts.

%% partial-label leanring
In the context of modeling noisy labels, our work is related to partial-label learning~\cite{ren2016label,cour2011learning,
nguyen2008classification} and multi-label multi-instance learning~\cite{surdeanu2012MIME}, which deals with the problem where each training instance is associated with a set of noisy candidate labels (where\textit{only one is correct}). Unlike these formulations, our \textit{joint} extraction problem deals with both \textit{classification with noisy labels} and \textit{modeling of entity-relation interactions}. In Sec~\ref{subsec:performance_comparison}, we compare our full-fledged model with its variants \textsc{CoType-EM} and \textsc{CoType-RM} to validate the Hypothesis on entity-relation interactions.

%################################
\begin{table}[t]
\begin{scriptsize}
\vspace{-0.2cm}
\hspace*{-0.3cm}
\begin{tabularx}{1.05\linewidth}{| m{1.15cm}|X|X|}
\hline
\textbf{Text}
%%% sentence 1
%& {A plane carrying the body of Polish First Lady \textit{\textbf{Maria Kaczynska}}, who was killed in an air crash in western \textit{\textbf{Russia}}, landed in Warsaw on Tuesday morning}
& {\textbf{\textit{Blake Edwards}}, a prolific \underline{filmmaker} who kept alive the tradition of slapstick \underline{comedy}, \underline{died} Wednesday of pneumonia at a hospital in \textbf{\textit{Santa Monica}}.}
%% sentence 3
& {\textit{\textbf{Anderson}} is survived by his wife Carol, \underline{sons} \textit{\textbf{Lee}} and Albert, daughter Shirley Englebrecht and nine grandchildren.}
\\ \hline
\textbf{MultiR}~\cite{hoffmann2011multiR}
%& \texttt{person}, {\color{red!70}\texttt{artist}}, {\color{red!70}\texttt{author}}
& $r^*$:~{\color{red!70}person:country\_of\_birth}, $\newline\Y_1^*$:~\{N/A\}, $\Y_2^*$:~\{N/A\}
& $r^*$:~{\color{red!70}None}, $\newline\Y_1^*$:~\{N/A\}, $\Y_2^*$:~\{N/A\}
\\ \hline
\textbf{Logistic}~\cite{mintz2009distant}
%& \texttt{person}, {\color{red!70}\texttt{artist}}, {\color{red!70}\texttt{musician}}
& $r^*$:~{\color{red!70}per:country\_of\_birth}, $\newline\Y_1^*$:~\{person\}, $\Y_2^*$:~\{{\color{red!70}country}\}
& $r^*$:~{\color{red!70}None}, $\Y_1^*$:~\{person\},$\newline\Y_2^*$:~\{person, {\color{red!70}politician}\}
\\ \hline
\textbf{CoType}
%& \texttt{person}, \texttt{politician}
& $r^*$:~person:place\_of\_death, $\newline\Y_1^*$:~\{person,artist,director\}, $\newline\Y_2^*$:~\{location, city\}
& $r^*$:~person:children, $\newline\Y_1^*$:~\{person\}, $\Y_2^*$:~\{person\}
\\ \hline
\end{tabularx}
\vspace{-0.2cm}
\caption{Example output of \textsc{CoType} and the compared methods on two news sentences from the \textbf{Wiki-KBP} dataset.}
\label{table:example_output}
\vspace{-0.2cm}
\end{scriptsize}
\end{table}
%################################

%\vspace{0.1cm}
\section{Conclusion}
\label{sec:conclusion}
%%% conclusion
This paper studies domain-independent, joint extraction of typed entities and relations in text with distant supervision. The proposed \textsc{CoType} framework runs domain-agnostic segmentation algorithm to mine entity mentions, and formulates the joint entity and relation mention typing problem as a global embedding problem. We design a noise-robust objective to faithfully model noisy type label, and capture the mutual dependencies between entity and relation. Experiment results demonstrate the effectiveness and robustness of \textsc{CoType} on text corpora of different domains. 
Interesting future work includes incorporating pseudo feedback idea~\cite{xu2013filling} to reduce false negative type labels in the training data, modeling type correlation in the given type hierarchy~\cite{ren2016label}, and performing type inference for test entity mention and relation mentions jointly. 
%\textsc{CoType} relies on minimal linguistic assumption (\ie, only POS-tagged corpus is required) and thus can be extended to different languages where pre-trained POS taggers in that language is available.  

\section{Acknowledgments}
Research was sponsored in part by the U.S. Army Research Lab. under Cooperative Agreement No. W911NF-09-2-0053 (NSCTA), National Science Foundation IIS-1017362, IIS-1320617, and IIS-1354329, HDTRA1-10-1-0120, and grant 1U54GM114838 awarded by NIGMS through funds provided by the trans-NIH Big Data to Knowledge (BD2K) initiative (www.bd2k.nih.gov). The views and conclusions contained in this paper are those of the authors and should not be interpreted as representing any funding agencies.

\newpage
\balance
\small
\bibliographystyle{abbrv}
\bibliography{17-cotype}

\end{document}